\documentclass[journal]{IEEEtran}
\usepackage{cite}

\ifCLASSINFOpdf
  \usepackage[pdftex]{graphicx}
\else
\fi
\usepackage[cmex10]{amsmath}
\usepackage{algorithm}
\usepackage{algpseudocode}

\usepackage{array}

 \usepackage[caption=false,font=normalsize,labelfont=sf,textfont=sf]{subfig}
\def\cross{\mbox{\boldmath $\times$}}	
\def\Trans{^{\rm T}}		
\def\cross{\mbox{\boldmath $\times$}}	
\def\Trans{^{\rm T}}		
\def\activityRange{{\mathcal I}}	
\def\motorSpace{{\mathcal M}}
\def\motorDim{m}
\def\motorSubspaces{{\hat{\motorSpace}}}

\def\sensorySpace{{\mathcal S}}
\def\sensoryDim{s}
\def\sensorySubspaces{{\hat{\sensorySpace}}}

\def\assocSpace{{\mathcal A}}
\def\assocDim{n}
\def\MSpoint{{\bf a}}

\def\facilField{D}

\def\facilRate{\eta_{\rm \facilField}}
\def\facilDecay{\tau_{\rm \facilField}}

\def\completionImage{G}
\def\assocRBFel{r}

\def\backwardVec{{\boldsymbol{\beta}}}

\def\backwardMatEl{B}
\def\backwardMat{{\bf \backwardMatEl}}
\def\backwardRate{\eta_{\rm \backwardMatEl}}
\def\backwardDecay{\tau_{\rm \backwardMatEl}}

\def\forwardMatEl{F}
\def\forwardMat{{\bf \forwardMatEl}}

\def\competMatEl{C}
\def\competMat{{\bf \competMatEl}}

\def\competVec{\boldsymbol{\chi}}

\def\delay{\delta t}
\def\completionRBF{{\boldsymbol{\gamma}}}

\newcommand{\Resolution}{\mathop{\mathrm{Resolution}}}         
\DeclareMathOperator*{\dprime}{\prime \prime}

\newcommand{\neqn}[2]{
\begin{equation}
#2
\label{eq:#1}
\end{equation}
}

\begin{document}
%
\title{Neurally Plausible Model of Robot Reaching Inspired by Infant Motor Babbling}
%
%
%

\author{Zahra~Mahoor,~
        Bruce~MacLennan,~\IEEEmembership{Member,~IEEE,}
        and~Allen~McBride
\thanks{Department of Electrical Engineering and Computer Science, University of Tennessee, Knoxville
        {\tt\small zmahoor@vols.utk.edu, maclennan@utk.edu, amcbri10@vols.utk.edu}}
\thanks{This paper is an extension of our previous work \cite{mahoor}.}
}

\maketitle

\begin{abstract}
In this paper we present a neurally plausible model of robot reaching inspired by human infant reaching that is based on embodied artificial intelligence, which emphasizes the importance of the sensory-motor interaction of an agent and the world. This model encompasses both learning sensory-motor correlations through motor babbling and also arm motion planning using spreading activation. This model is organized in three layers of neural maps with parallel structures representing the same sensory-motor space. The motor babbling period shapes the structure of the three neural maps as well as the connections within and between them. We describe an implementation of this model and an investigation of this implementation using a simple reaching task on a humanoid robot. The robot has learned successfully to plan reaching motions from a test set with high accuracy and smoothness.
\end{abstract}


\ifCLASSOPTIONpeerreview
 \begin{center} \bfseries EDICS Category: 3-BBND \end{center}
\fi
%
\IEEEpeerreviewmaketitle

\section{Introduction}
%
%
%
%
\IEEEPARstart{A}{genuinely} autonomous robot must adapt to unforeseen circumstances and learn from past encounters. Such adaptation is a challenge for traditional robot designs, which are best suited to narrowly-defined tasks and environments \cite{Law2011a}. One approach to designing a robot that continues to adapt to novel relationships between its body and environment is to provide it with a mechanism to investigate these relationships.

In contrast to robots, human infants quickly gain sensory-motor competence that they can generalize to diverse situations, even as their bodies change drastically. In particular, limb coordination is learned in a few months, as part of Piaget's first stage of infant development.
In developing sensory-motor skills through exploration, infants gain an internal representation of their bodies, which they can use to interpret interactions with their environment \cite{Piek}. We take inspiration from infant development to build a computational model that is: (1) {\it embodied}, (2) {\it developmental}, and (3) {\it neurally-plausible}.

An embodied model of sensory-motor coordination has the advantage that much of the information about the physical constraints of the body-environment interaction remains implicit in the environment itself, making both the learning and execution of the model more efficient.
A developmental model allows for a control mechanism that has less prior understanding of body-environment interactions and that can adapt better to unforeseen changes in the environment and even of the robot body itself (\textit{e.g.}, a replacement arm incorporating novel technology).
Finally, by being neurally-plausible, at least in its broad strokes, our model and its future refinements may capture some of the design work already performed by evolution and provide useful hypotheses back to neuroscience.

We will use the term {\em motor-sensory} as an intentional inversion of the usual ``sensory-motor,'' since the latter can suggest an input-compute-output information-processing model of cognition, which is inadequate. Rather, perception is an active process, which typically depends on the motion of the agent in its environment \cite{gib:79}. 
This is critical because the motor activity directs the gathering of perception of the environment in a more fruitful way \cite{spo:04}.
``Motor-sensory'' reminds us that motor activity is fundamentally prior to sensation and perception, although of course the two occur in a tight loop established by the agent's continuous physical engagement with its environment. The physical constraints of body-environment interactions define manifolds of possible trajectories in this motor-sensory space \cite{kun:04}.
In our model, a {\em motor-sensory space} defines the interface between the agent's internal control processes (e.g., its nervous system) and the body-environment system. This space is defined by all the sensory inputs and all the motor outputs of the nervous system or corresponding artificial control system. For all but the simplest animals this is a very high-dimensional space. 

\section{Related Work}
The present work is an instance of developmental robotics because an agent learns incrementally  motor-sensory coordination and prediction through self-exploration \cite{Lungarella2003,Asada2009}. It is neurally plausible, uses motor babbling to learn inverse dynamics for arm trajectories, and is tested on a real robot. Here we review related studies that serve as a basis for this paper. Each of these studies shares some, but not all, of the qualities listed above that make our work an original contribution to the field.

\cite{Gomez2004a} mimics infant development in robot reaching in three concurrent stages but the focus is not the neural plausibility of the reaching model.
\cite{Caligiore2008} uses motor babbling and Hebbian learning to form motor-sensory correlations for reaching and grasping an object in the presence of an obstacle. Some parts of their method, such as learning the parameters of their pattern generator, or inverting the kinematic process, are not neurally plausible.

In \cite{Asuni2005}, a neural model was built for visuomotor coordination of a robotic manipulator in the reaching task. This model uses a self-organizing neural network to learn the correlation of motor actions and sensory feedback. This system maps between the position of the arm in the 3D Cartesian space and its joint space (inverse kinematic). 
In \cite{Demiris2005}, both real and simulated robots learn multiple forward models without having any prior information using motor babbling and a Bayesian belief network. In this system, which is inspired by human hand movement, an association between motor commands and the position of the moving gripper is learned (inverse kinematic).

In \cite{Lee2007a}, a robotic system learns the correlation between proprioceptive and motor space by taking advantage of natural constraints. Those natural constraints are active and inactive sensing, use of objects, and sensory resolution. 

In \cite{Lee2007a, Lee2007b,Law2011a }, the LCAS (Lift-Constraint, Act, Saturate) algorithm was introduced to learn hand/eye coordination. At the beginning of the LCAS cycle, all or almost all constraints are imposed, and there is little room for complex activity. In each cycle, the system gradually removes a restriction and explores (Act) all the possible new experiences until the learning saturates. 
The computational framework of this algorithm is based on a two-dimensional map, where the map consists of circular overlapping and regularly spaced receptive fields. In this work, a correlation map between motor and sensor space is built, and it doesn't focus on trajectory planning and reaching. 
 
In \cite{Saegusa2008a}, a learning system was developed to predict future sensor values from current sensor values and motor commands. The motor-sensory learning procedure is divided into two stages of exploration and learning. The system alternates between these two stages until the desired performance is reached. The exploration strategy is improved in \cite{Saegusa2009IEEE/RSJ} and \cite{Saegusa2009HSI}. This system learns motor-sensory prediction rather a trajectory planning. 


\cite{Laschi2008} offered a predictive motor-sensory coordination system inspired by infant development for robot reaching using neuro-fuzzy networks. In this work reaching controls the final position and orientation of the arm end effector, but not the arm's trajectory.

\cite{Rolf2010a} uses goal babbling, as opposed to motor babbling, as a strategy to learn inverse kinematics in reaching. Since motor babbling focuses on the exploration of the entire joint space, goal-directed babbling is offered as a feasible alternative exploration for arms with many degrees of freedom. In contrast, we are trying to solve the inverse dynamics problem. The robustness of the goal-babbling approach was tested in \cite{Rolf2010b} for body growth both on a simulated robot arm and on the iCub humanoid robot.

\cite{Dewolf2011} offers a neurally plausible approach for motor control of reaching using optimal feedback control. Functions of this model are mapped to parts of the brain that are known to be involved in motor control. But their approach was not tested on any real robotic system, and the model of reaching is not inspired by infants. 
\cite{Law2014b} is an excellent example of a longitudinal approach to development that starts from motor babbling and continues to the reaching and grasping stage. Simulated motor and sensory spaces are represented by overlapping maps of fields that resemble topographic maps in the brain. The motor-sensory correlation is stored in the links that connect fields of the corresponding maps. The focus of this work is capturing the developmental stages of reaching and not the neural plausibility of the model. 
\cite{Caligiore2014} introduced a computational model for the development of reaching by integrating reinforcement learning, equilibrium points, and minimum variance. The focus of this work is capturing the essential features of reaching and not the neural plausibility of the model.

In the following section, we describe our proposed conceptual model of reaching. In section \ref{neural-model}, we explain a neural model for the proposed abstract model. In section \ref{model-implement} we explain the implementation of this model. In section \ref{experiments} we present our experiments using a humanoid robot and finally discuss our results from experiments in section \ref{discussion}. 

 
\section{CONCEPTUAL MODEL}
\subsection{Motor-sensory Phase-space and Trajectory Bundles}
\label{sec:Formation of Trajectory Bundles}
We take our inspiration from the embodied development of the human motor-sensory system, in which an infant must learn the dynamical relationship between its body and environment. Focusing on the arm, we introduce a model for learning the correlation between motor action and consequent sensation.


Let the space $\sensorySpace$ represents the possible states of the sensory input.
If there are $\sensoryDim$ sensor inputs, and if for convenience we normalize them to $\activityRange = [-1,1]$, then $\sensorySpace = \activityRange^\sensoryDim$.
However, the sensory space is divided into $\sensorySubspaces$ disjoint subspaces
$\sensorySpace = 
\sensorySpace_1 \cross \sensorySpace_2 \cross \cdots
 \cross \sensorySpace_\sensorySubspaces$
of dimension $\sensoryDim_1, \sensoryDim_2, \ldots, \sensoryDim_\sensorySubspaces$,
respectively.
These subspaces correspond to distinct sensory modalities; 
for example $\sensorySpace_1$ might be haptic input,
$\sensorySpace_2$ might be proprioceptive input,
and $\sensorySpace_3$ might be visual input.

The space $\motorSpace = \activityRange^\motorDim$ represents possible states of the motor output system, which, like the sensory system, comprises disjoint systems
$\motorSpace = \motorSpace_1 \cross \motorSpace_2 \cross \cdots \cross \motorSpace_\motorSubspaces$.
For example, one such subspace might represent the muscles or actuators of the fingers of the hand.

In this model, we are generally concerned with trajectories in motor-sensory space, $\assocSpace = \motorSpace \cross \sensorySpace$,
which has dimension $\assocDim = \motorDim \sensoryDim$.
Since neurons represent values in $\activityRange$ with low precision (about 0.1), the space $\assocSpace = \activityRange^\assocDim$ may be characterized as a space of small size (diameter)
but very high dimension.

\begin{figure}[!t]
  \centering
  \framebox{\parbox{2in}{\includegraphics[scale=0.5]{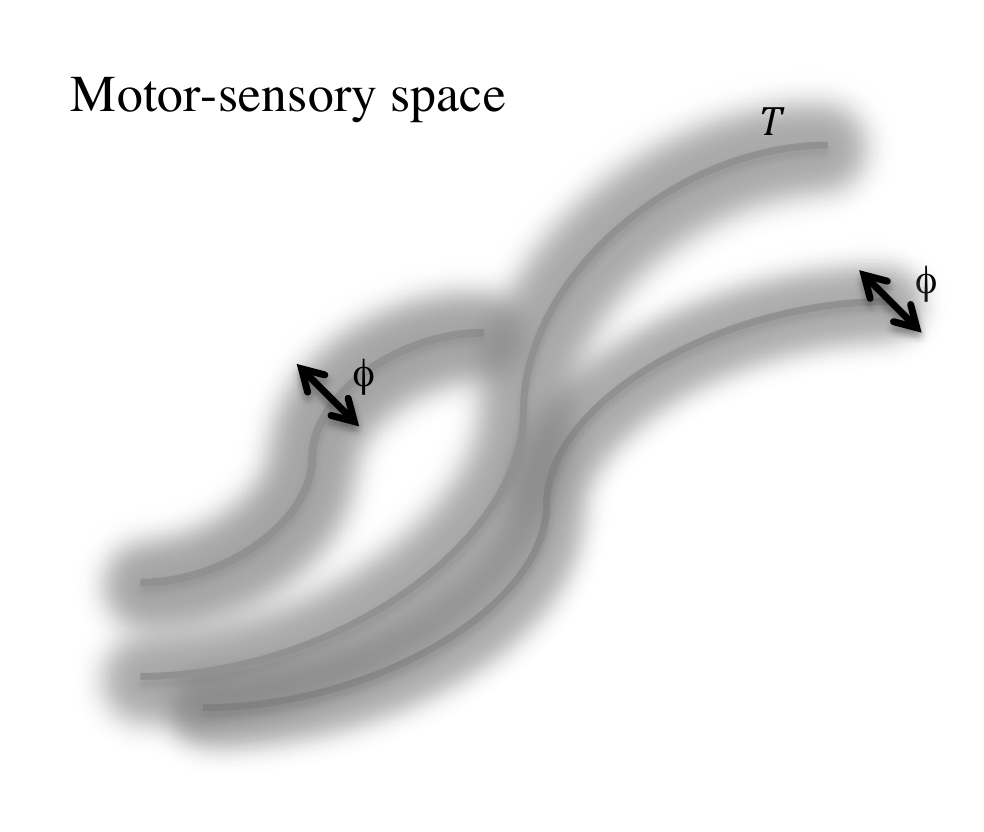}}}
  \caption{A conceptual bundle with three trajectories where $\phi$ represents the width of trajectories to fuzzily the dynamic.}
  \label{fig:abs}
\end{figure}

A major component of this approach is learning correlations between motor actions and consequent sensations, for $\MSpoint \in \assocSpace$. The goal is to find regions of $\assocSpace$ that are dynamically feasible. One way to construct this field is by recording motor-sensory trajectories through $\assocSpace$, thus constructing trajectory bundles. Let $T$ be a motor-sensory trajectory which is a sequence of points in $\assocSpace$, we consider a ball around each point along the trajectory as it carves $\assocSpace$ space. This ``fuzzifies'' the trajectory, reflecting the fact that the dynamics are continuous. Fig.~\ref{fig:abs} shows a conceptual bundle of three trajectories in which parameter $\phi$ is used to show the fuzziness of trajectories. 

Space $\assocSpace$ has very high dimension, and direct neural implementation of this correlation learning could be computationally impractical. In our model, we use dimension reduction to create a more computationally tractable space $\assocSpace^\prime$ for learning correlations. There is evidence of dimension reduction mechanism in several brain areas, \textit{e.g.} cerebellum and other motor-sensory systems. 

\subsection{Conceptual Trajectory Planning and Execution}
The abstract path planning process conjectures a trajectory through the abstract motor-sensory phase space $\assocSpace$ to reach a dynamical goal from a dynamical starting point. The goal is represented by an ``image of completion'' $\completionImage : \assocSpace \rightarrow \activityRange$,
which measures the attractiveness of motor-sensory goal states.
For example, in the case of an infant reaching for and grasping an object, the image of completion is the perception of grasping the object (tactile, proprioceptive, visual, etc.).
Such an image of completion might be elicited by the sight of an interesting object at a particular place in the infant's visual field.
The visual information provided about the object's location and properties (size, material, etc.) combines with the infant's goal (holding it) to generate the image of completion.
For example, the desire to grasp the object might generate goal haptic inputs in $\sensorySpace_1$, and the perceived location of the object would generate goal activity defined over the proprioceptive and visual fields (e.g., $\sensorySpace_2$ and $\sensorySpace_3$).
The purpose of the path planning process, then, is to find an abstract trajectory from the current motor-sensory state into the goal region.

The process can be described intuitively as follows.
The trajectory bundles defines feasible ``paths'' through $\assocSpace$, analogous to ant trails, but in a very high-dimensional phase space.
Or, to invert the metaphor, we can think of space outside the trajectory bundles as representing impassible regions through $\assocSpace$. Then the path planning process can be envisioned as spreading activity from $\completionImage$ until it reaches the current motor-sensory state.
Path execution is implemented by following the gradient of this signal from the current location in $\assocSpace$ to the goal region.
However, it is better if the motor-sensory state changes only if the path signal is above a certain threshold.

This simple mechanism has several desirable properties.
First, the motor-sensory system will not begin to seek the goal until a feasible dynamical strategy is determined.
Second, the path to the goal will tend to be good (in terms of facility and length in motor-sensory space), although not necessarily optimal.
Third if the goal state changes or disappears, the trajectory will change, either to seek the new goal or to cease seeking a goal (if the signal drops below threshold).
Finally, and most importantly, if there are any perturbations of the trajectory, for example from unexpected sensory input, then the process will automatically follow the best available above-threshold path in its new state.

While it is convenient to describe the path following process in terms of the gradient, the actual neural mechanisms will be only an approximation. 
For example, if there are two or more equally attractive paths, inherent stochastic mechanisms will cause it to break the symmetry and pick one of them. Thus the process will not be stalled by saddle points or split the difference between equally attractive paths.

\section{NEURAL MODEL}
\label{neural-model}

\subsection{Neurally-Plausible Bundle Formation}
\begin{figure}[!t]
\centering
\framebox{\parbox{3.4in}{\includegraphics[scale=0.5]{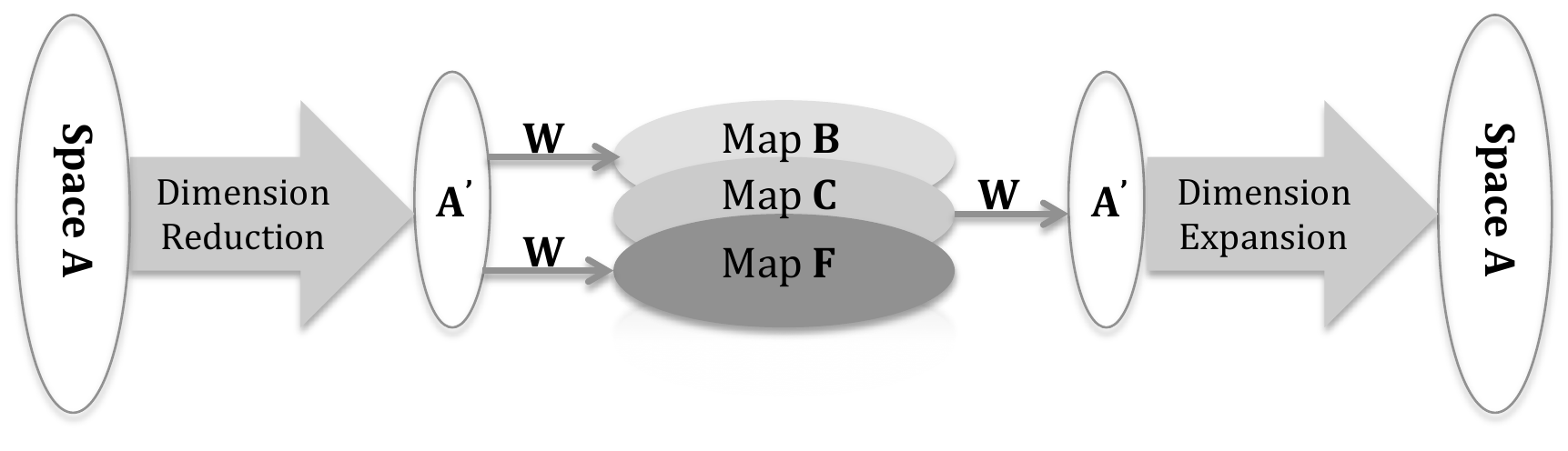}}}
\caption{An overview of our reaching model with three neural maps at its core and dimension reduction and dimension expansion modules.}
\label{fig:overview}
\end{figure}

\begin{figure}[!t]
  \centering
 \framebox{\parbox{2in}{ \includegraphics[scale=0.5]{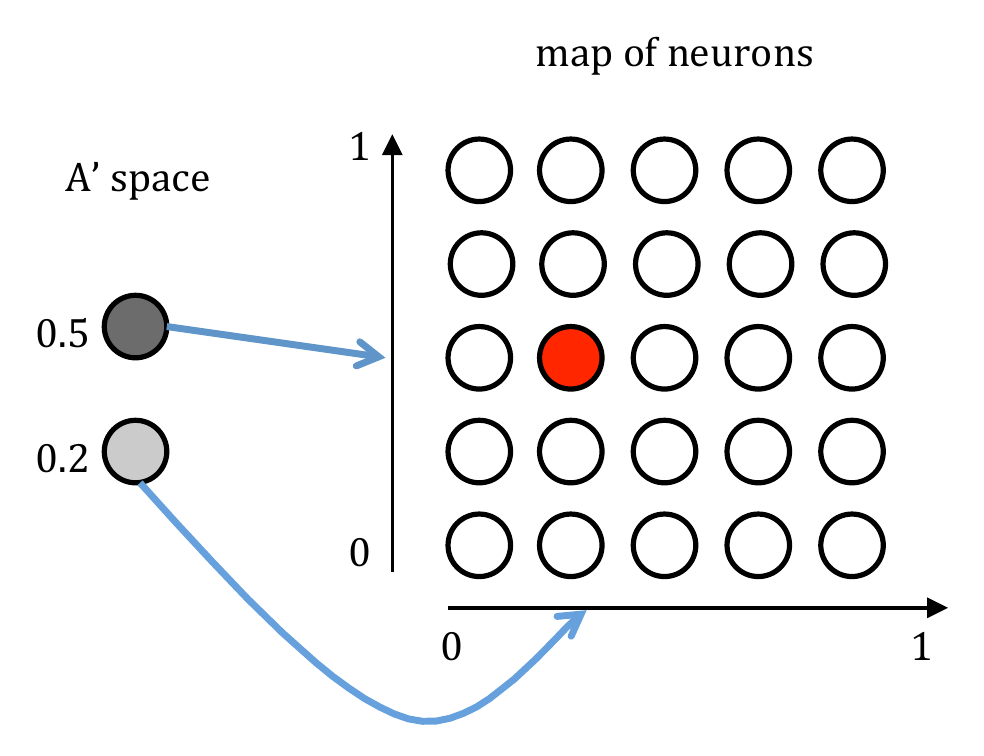}}}
  \caption{We conceive each neural map as an array, with each dimension of the array representing one of the neurons in $\assocSpace^\prime$. In this example, $\assocSpace^\prime$ has two neurons so the neural map is a 2D array. The location of an activated neuron in a map represents a vector of activation levels for $\assocSpace^\prime$. The map neuron's location with respect to the first dimension of the array is the activation level of the first neuron in $\assocSpace^\prime$, and so on.}
\label{fig:map}
\end{figure}

\begin{figure}[!t]
\centering
\framebox{\parbox{2.1in}{\includegraphics[scale=0.43]{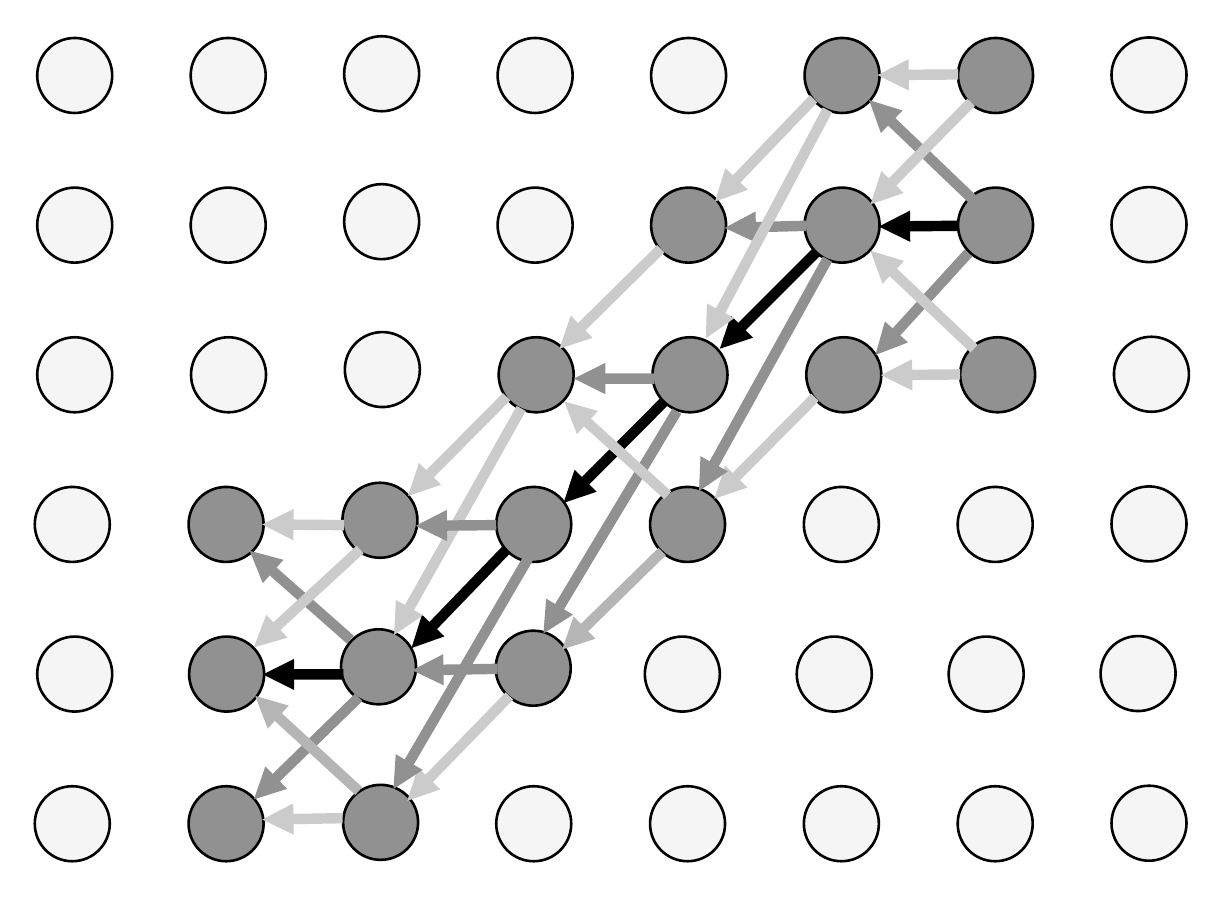}}}
\caption{A simplified neural representation of a trajectory bundle in the map $\backwardMat$ where the gray nodes are inside the bundle and color of the edges represent strength of the connections among the neurons in the bundle. Darker edges show the stronger connections.}
\label{fig:bundle-edges}
\end{figure}

The central feature of our model is the encoding of trajectory bundles in three maps of neurons with a parallel structure representing the same motor-sensory space. We refer to them as the \textit{backward}, \textit{forward} and \textit{competition} maps. Fig.\ \ref{fig:overview} shows an overview of this model with the three neural maps in the center and the dimension reduction and dimension expansion modules on the sides. A particular motor-sensory state is represented by localized activity over the maps shown in Fig.\ \ref{fig:map}. In addition, trajectories are defined by shifting activities among neurons with overlapping receptive fields. 
The connections between successively activated neurons encode both reverse-time correlations for path planning and forward-time correlations for path execution. The backward map $\backwardMat$ represents connections from neurons activated at time $t+\delay$ to neurons activated at time $t$ and the connection strength to neuron $i$ from neuron $j$ is given by
\neqn{dot-backwardMat}{
 \dot{\backwardMatEl}_{ij} =
 \facilRate (1-\backwardMatEl_{ij}) \assocRBFel_i(t) \assocRBFel_j(t+\delay)
  - \backwardMatEl_{ij} / \facilDecay
,}
where $\assocRBFel_k$ is the activation of $k$-th neuron of the neural map. The learning rate $\facilRate$ is small so that trajectory bundles evolve slowly. To allow these connections to adapt to changes in body dynamics (e.g., due to growth), a slow 
decay term $\facilDecay$ is added to this equation. Fig.\ \ref{fig:bundle-edges} illustrates a simplified neural representation of map $\backwardMat$ with neurons and connections among the neurons in a bundle. The connections are stronger in the center of the bundle compared to the connections in the sides. 

Connections between the forward map $\forwardMat$ and the competition map $\competMat$ represent forward connections for path execution from neurons activated at time $t$ to neurons activated at time $t+\delay$. These connections evolve as, 
 \neqn{dot-forwardMat}{
 \dot{\forwardMatEl}_{ij} = 
 \facilRate (1-\forwardMatEl_{ij}) \assocRBFel_i(t+\delay) \assocRBFel_j(t)
  - \forwardMatEl_{ij} / \facilDecay .}
Hence, $\forwardMat = \backwardMat\Trans$. The connections in both forward map and the backward map are only added among neurons that represent nearby points in the space $\assocSpace^\prime$ and no other neurons.
Finally, in the competition map $\competMat$, mutually inhibitory connections between nearby neurons implement a competitive network. Neurons in these maps are Radial Basis Functions (RBF) in which centers are determined by a neural weight matrix $\mathbf{W}$. Connections in $\mathbf{W}$ represent the receptive fields of neurons in maps $\backwardMat$ and $\forwardMat$ from space $\assocSpace^\prime$ and are normalized ($||\mathbf{W}_{i}||=1$). In addition, normalized connections $\mathbf{W}$ also represent the projection fields of neurons in map $\competMat$ to space $\assocSpace^\prime$. Because the vectors comprising $\mathbf{W}$ are normalized, activity levels of neurons in the neural maps are inversely proportional to the Euclidean distance between the centers represented by those neurons and a given point in space $\assocSpace^\prime$. We consider the weights $\mathbf{W}$, as well as the underlying topology of the neural maps, to represent the result not of motor babbling itself but of prior development as determined by evolution or other developmental processes.

\subsection{Neurally-Plausible Trajectory Planning and Execution}
\begin{figure}[!t]
\centering
\framebox{\parbox{3.3in}{\includegraphics[scale=0.5]{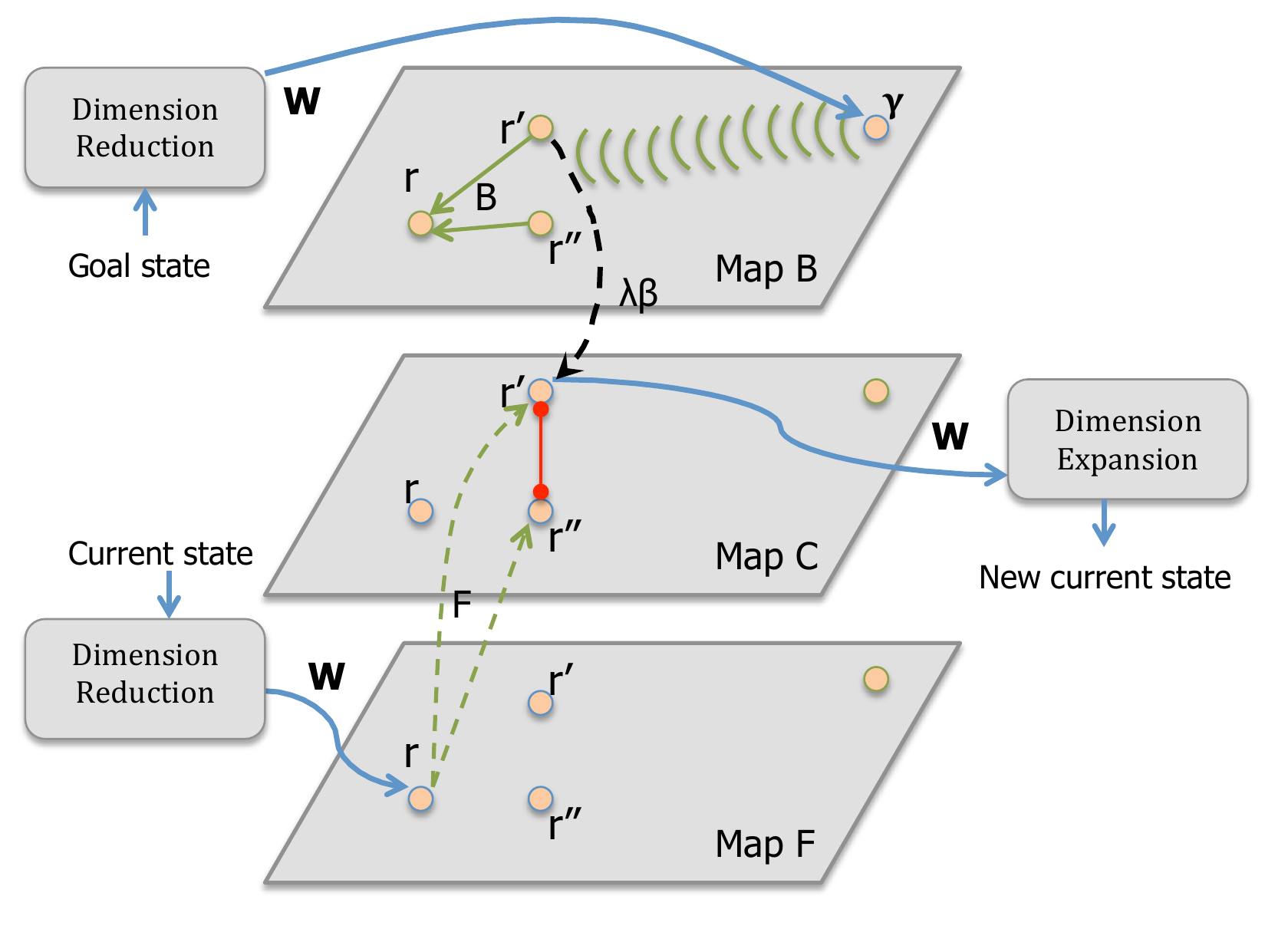}}}
\caption{Neural architecture for implementing the path planning and execution process. Activity in map $\backwardMat$ spreads from the goal state $\completionRBF$, and the current state $r$ in map $\forwardMat$ excites nearby neurons $r^{\prime}$ and $r^{\dprime}$ in map $\competMat$. Competition among excited neurons in map $\competMat$ leads to firing of neuron $r^{\prime}$, which represents the new motor-sensory state.}
\label{fig:neural-arch}
\end{figure}

After trajectory bundles are created, an agent can find a trajectory or path through the abstract motor-sensory phase space $\assocSpace$ from a dynamical starting point to a goal. The goal or image of completion initiates the process of path planning. Activity spreads through the backward connections in the neural map $\backwardMat$ outward from the goal, that is, from the neurons that participate in the image of completion (see Fig.\ \ref{fig:neural-arch}).
Let $\completionRBF$ be the activity of goal neurons and $\backwardVec$ be the backward spreading activation. Then the dynamics of backward activation $\backwardVec$ is:
\neqn{eq:dot-backwardVec}{
\dot{\backwardVec} =
 \backwardRate (\backwardMat \backwardVec + \completionRBF)(1-\backwardVec)
  - \backwardVec / \backwardDecay ,
}
This update rule implements spreading activation, weighted according to the synapse weights in $\backwardMat$. Here, $\backwardRate$ represents the activation rate. A decay term $\backwardVec / \backwardDecay$ is included so that if the goal changes, the potential paths will quickly readjust. 

Path execution begins when neurons in map $\competMat$ receive input from neurons in map $\backwardMat$ (representing path planning) as well as from neurons in map $\forwardMat$ (representing the current motor-sensory state). Activity in $\backwardMat$ activates corresponding neurons in map $\competMat$ to a degree of $\lambda \backwardVec$. At the same time, activity of neuron $r$ in $\forwardMat$ activates potential successor neurons $r^{\prime}$ and $r^{\dprime}$ in map $\competMat$. Activated neurons in map $\competMat$ compete to define the next state of the planning; the neuron $r^{\prime}$ that was maximally excited by both the current state $r$ and the backward connections from the goal state is the winning neuron. This neuron fires and defines the next motor-sensory state in $\assocSpace^\prime$. This state is translated back from $\assocSpace^\prime$ to $\assocSpace$ to generate both motor signals and sensor prediction. The winning neuron shifts to a refractory state for the rest of the planning and execution; this refractory state helps to avoid cycles in the path planning. 

\section{MODEL IMPLEMENTATION}
\label{model-implement}
We use trajectories resulting from motor babbling in three passes. The first pass trains a dimension reduction module. The second pass structures the three parallel neural maps $\forwardMat$, $\backwardMat$, and $\competMat$. The third pass determines the weights of synapses within and among these neural maps. In the following sections we explain the implementation of our model, which includes a deep autoencoder for the purpose of dimension reduction, neural representation of phase-space, trajectory bundle formation, and path planning and execution processes. 

\subsection{Reduction of Dimension}
Autoencoders are simple neural networks that are used to transform inputs into outputs with the least possible amount of loss. The goal is to make the output the same as the input in a network with a central bottleneck. Autoencoders use backpropagation to find synapse weights that encode the input in the middle layer \cite{autoencoder2006}. Backpropagation is a form of error-driven learning that can be implemented using a neurally plausible model as proposed by \cite{CCNbook}.

Both inputs and outputs of the autoencoder are points in the motor-sensory space $\assocSpace$, and the bottleneck represents the same points in the reduced space $\assocSpace^{\prime}$. After training, the front and back halves of the autoencoder serve as an encoder and decoder respectively. The encoder module is used to project motor-sensory states from space $\assocSpace$ to the reduced-dimensional space $\assocSpace^{\prime}$. Similarly, the decoder is used to transform states from space $\assocSpace^{\prime} $ to space $\assocSpace$.

\subsection{Representation of Motor-sensory Space}
\label{sec:neural-map}
After transforming the babbling trajectories into a lower dimensional space, we create a set of \textit{RBFs} that is the same for neural maps $\forwardMat$, $\backwardMat$, and $\competMat$. Consider $T=\{\MSpoint_{1}, .., \MSpoint_{i}, \MSpoint_{i+1},.., \MSpoint_{f}\}$, $\MSpoint_{i} \in \assocSpace$, a trajectory in the high-dimensional phase-space $\assocSpace$, and $T^{\prime} = \{\MSpoint^{\prime}_{1}, .., \MSpoint^{\prime}_{i}, \MSpoint^{\prime}_{i+1},..,\MSpoint^{\prime}_{f}\}$, $\MSpoint^{\prime}_{i} \in \assocSpace^\prime$, the same trajectory in the reduced space $\assocSpace^{\prime}$. The $j$th feature of $\MSpoint^{\prime}_{i}$ is shown by $a^{\prime}_{ij}$. We want to create a matrix $\mathbf{W}$ which serves as the receptive fields of neurons in neural maps $\forwardMat$ and $\backwardMat$, and the projection field of neurons in map $\competMat$. To make each weight vector of $\mathbf{W}$ normalized, ($||\mathbf{W}_{i}||=1$), we calculate an additional pseudo-feature based on the features in $\assocSpace^{\prime}$ such that this pseudo-feature guarantees $||\mathbf{W}_{i}||=1$. This means the extra feature called $W_{i1}$ can be calculated simply by the other features as, 

\neqn{pseudo-feature}{W_{i1}=\sqrt{1-\sum_{j=2}^{|\assocSpace^\prime|}W_{ij}^2}}

Assume we have a function \textit{Resolution} that takes a vector of values for a given feature and returns a desired resolution for that feature. Then the resolution of each feature is given by
\neqn{res}
{{\rm res}_{j} = \Resolution(\{a^{\prime}_{ij}| i=1:k \}), \mbox{ for } j=1:|\assocSpace^\prime| ,}
where $k$ is the number of points in all babbling trajectories. Any function returning a set of intervals covering a feature's range can be used, and we discuss one example in section \ref{experiments}.

High-resolution maps assure a smooth motion trajectory by activating different neurons for different motor-sensory points in $T^{\prime}$. In order to efficiently store such fine-grained maps of motor-sensory space, we store only those neurons representing points in or near the trajectory bundles resulting from babbling. Each motor-sensory point $\MSpoint^{\prime}_{i}$ of the babbling data as well as each of its neighboring points as determined by function \textit{Resolution} is assigned as the center of an RBF.

\subsection{Implementation of Trajectory Bundle Formation}
Alg.\ \ref{BundleLearning} describes the bundle formation process (following eq.~\ref{eq:dot-backwardMat} and eq.~\ref{eq:dot-forwardMat}) that occurs in maps $\backwardMat$ and $\forwardMat$ through motor babbling. A babbling trajectory in the reduced space is passed as an input to this procedure. Iteratively, points along the trajectory fire a set of neurons from the neural map $\backwardMat$, and reverse-time connections between firing neurons are strengthened. The function \textit{find\_firing\_neurons} defines the top $\phi$ neurons fired for a motor-sensory point along the babbling trajectory. At iteration $i+1$, fired neurons have their connections to the previously fired neurons from iteration $i$ increased by weight $w$. In this procedure, $\phi$ stands for the width of bundles. By setting $\phi$ to a value larger than one, we can create synapses not only between the maximally firing neurons but also between neighboring neurons with a lesser level of activity. In Alg.\ \ref{BundleLearning}, the function \textit{calculate\_weight} determines the strength of connections as a linearly decreasing function of distance from the middle of the bundles. The \textit{update} function increases the old connection's strength by $w$ and guarantees that the strength of connections is not above $1.0$. After successfully creating map $\backwardMat$, we copy the connections from map $\backwardMat$ into map $\forwardMat$ with reversed direction. 

\begin{algorithm}[H]
\caption{Trajectory Bundle Formation.} 
\label{BundleLearning}
\begin{algorithmic}[1]
\Procedure{BundleFormation}{${\rm neural\_map}, T',\newline \phi, \facilDecay,\facilRate$}
\State $n \gets$ find\_firing\_neurons$(T^{\prime}[1],\phi)$
\For{$k\leftarrow 2 $ to $|T^{\prime}|$}\Comment{for all points along $T^{\prime}$}
	\State $m \gets$ find\_firing\_neurons$(T^{\prime}[k],\phi)$
	\For{$i$ in $n$ and $j$ in $m$} 
		\State $w \gets$ calculate\_weight$(i,j)$
	        \State update$(\backwardMat[j,i], w, \facilRate, \facilDecay)$
	\EndFor
  \State $n \gets m$
  \EndFor
  \State $\forwardMat \gets \backwardMat^T$
  \State \textbf{return} neural\_map
\EndProcedure
\end{algorithmic}
\end{algorithm}

\subsection{Implementation of Path Planning and Execution}
\label{PathPlanningImp}
Phase-space trajectories are represented by changing patterns of activity over the neurons in the neural maps. Trajectory planning occurs in the reduced space $\assocSpace^\prime$. Alg. \ref{PathPlanning} describes the implementation of path planning and execution. 
In this procedure, $\backwardVec$ initially is set to zero for all the neurons in map $\backwardMat$. We iteratively update $\backwardVec$ until the end of path execution or for a certain number of steps, \textit{max\_step}. Meanwhile, $\competVec$ (activity of neurons in map $\competMat$) is updated for the neighbors of current state $r$ in map $\competMat$ where the first term, $\lambda \backwardVec$, reflects the weight of connections from map $\backwardMat$ and the second part, $\forwardMat[r, n]$, reflects the weight of forward connections from map $\forwardMat$. The competition between nearby neurons in map $\competMat$ is computed by argmax. The next current state $r^{\prime}$ is added to list \textit{fired\_neurons}, which keeps track of neurons that have been fired throughout path execution and are in their refractory state. The \textit{transform} function projects the motor-sensory state represented by $r$ from $\assocSpace^\prime$ back to $\assocSpace$, from which motor commands can be sent to the arm for execution.

\begin{algorithm}
\caption{Path Planning and Execution.}
\label{PathPlanning}
\begin{algorithmic}[1]
\Procedure{PathPlanning}{neural\_map, start, goal,\newline $\backwardRate, \backwardDecay, \lambda$,  max\_step}
   \State $\backwardVec \gets 0.0$ for all neurons in neural\_map
   \State  $\competVec \gets 0.0$ for all neurons in neural\_map

   \State $r \gets$ start \Comment{current state}
   \State fired\_neurons $\gets \{r\}$
   \State step $\gets 0$
   \While{$ r \neq$ goal and step $<$ max\_step}
       \For {each $n$ in neural\_map}
          \State $\backwardVec \gets
          \backwardVec + \backwardRate (\backwardMat \backwardVec + \completionRBF)(1-\backwardVec)- \backwardVec / \backwardDecay$
          \If{$\backwardVec >0$ and $\forwardMat[r,n]>0$}
              \If{$n$ is not in fired\_neurons}
           	\State $\competVec \gets \lambda \backwardVec + \forwardMat[r,n]$
             \EndIf
           \EndIf
        \EndFor
    \If{$\max(\competVec)>0$}
        \State $r^{\prime} \gets \mathop{\rm argmax}(\competVec)$\Comment{winning neuron, $r^{\prime}$}
        \State $r \gets r^{\prime}$\Comment{new current state}
        \State (motor-sensory) $\gets$ transform$(W[r])$
        \State add  $r$ to fired\_neurons
    \EndIf
    \State step $\gets$ step + 1
    \State $\competVec \gets 0.0$ for all nodes
   \EndWhile\label{pathwhile}
  \EndProcedure
\end{algorithmic}
\end{algorithm}

\begin{figure}[!t]
  \centering
  \subfloat[]{\includegraphics[scale=0.16]{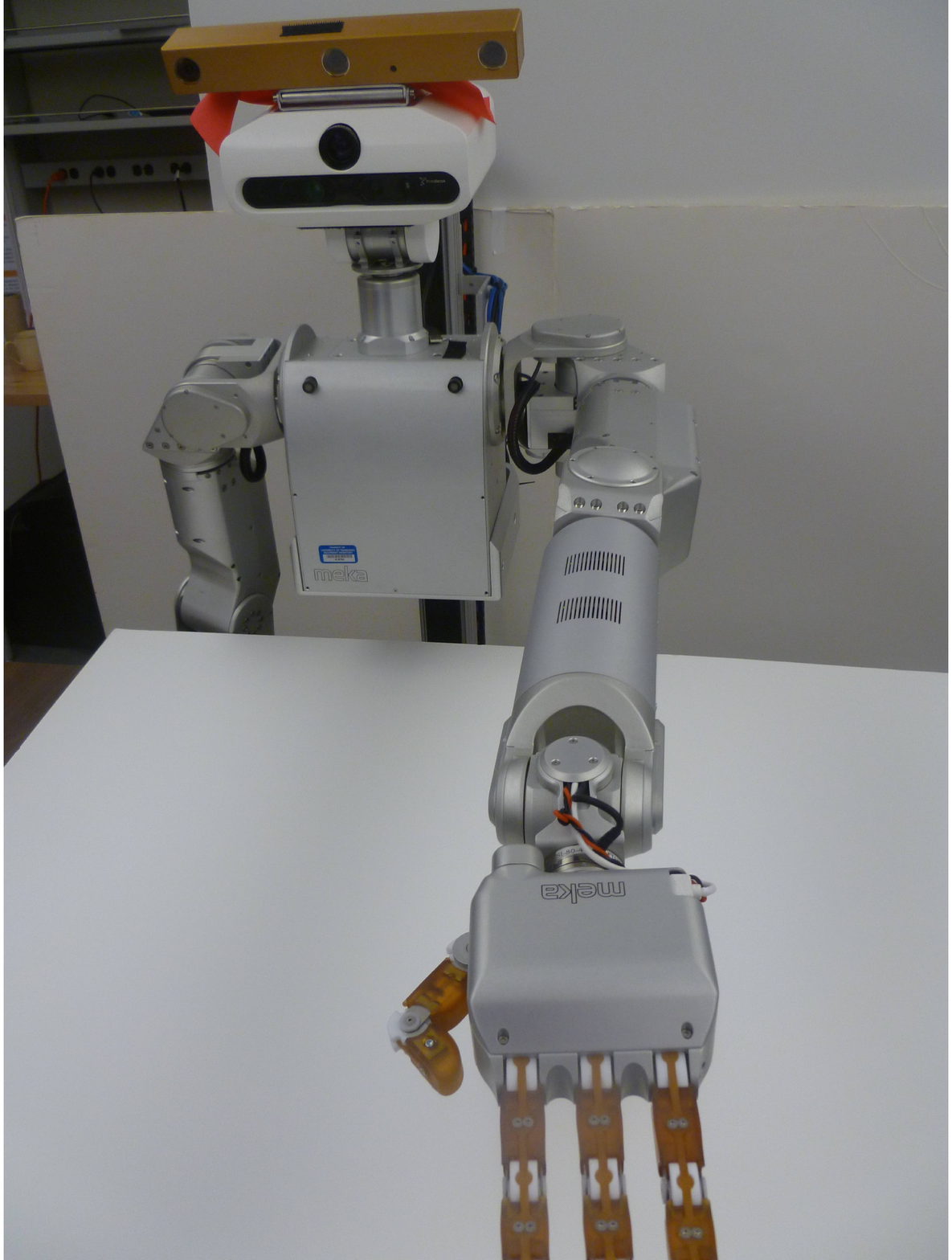}\label{fig:rosie}}\quad
  \subfloat[]{\includegraphics[scale=0.5]{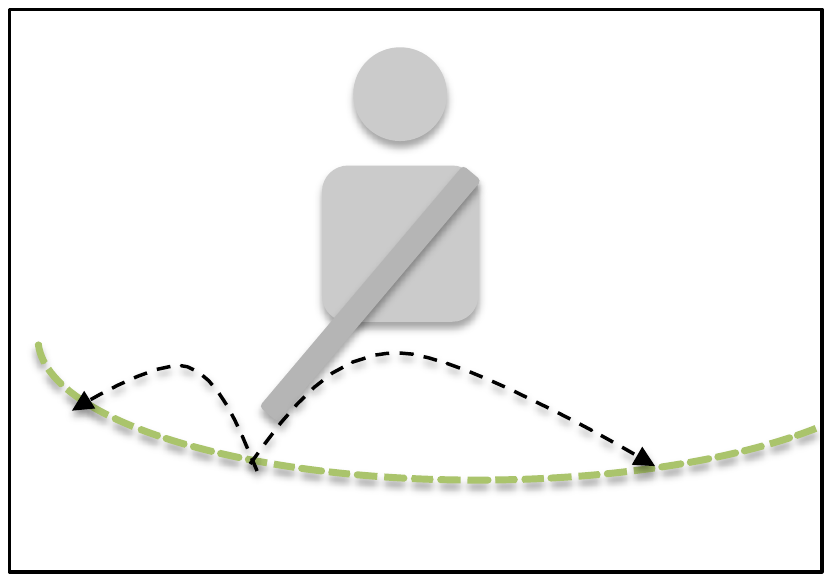} \label{fig:arc}}
  \caption{Experimental settings: 
  (a) Meka Robotics M3 mobile humanoid robot ``Rosie.''
  (b) The humanoid robot randomly explores an arc in front of its left arm with a fixed start pose. The initial and final positions of babbling trajectories are on this arc, but there is no constraint on points between the initial and final positions.}
\end{figure}

\section{EXPERIMENTS}
\label{experiments}
We implemented our algorithm with a Meka Robotics M3 mobile humanoid robot named Rosie (Fig.~\ref{fig:rosie}). Rosie has a 0-DOF torso and two 7-DOF arms, each of which has a 6-DOF end effector (hand) with four fingers. We used 5 joints in Rosie's left arm (3 joints in the shoulder and 2 joints in the elbow) for our experiment. Random goal positions were generated in an arc in front of the robot for training (babbling) and testing our reaching controller in 3D space. All communication with the robot is through the Robot Operating System (ROS). For training, trajectories are generated with MoveIt! \cite{Sucan2014}. Motor commands consist of joint positions and velocities.

In the first pass through the training trajectories, we trained an autoencoder using backpropagation with the $\tanh$ activation function. Before backpropagation, the training data was normalized to the range $[0,1]$ by dividing each feature by its maximum value and subtracting its minimum value. The learning rate was $0.1$ and the number of epochs was $10^{6}$. In the second pass of training, neural maps were constructed by calculating a desired resolution for each feature (section \ref{sec:neural-map}). This resolution was set to the median distance of consecutive points along all trajectories in space $\assocSpace^\prime$. The third pass through the babbling trajectories was to make connections among neurons in the neural map based on Alg.\ \ref{BundleLearning}. 

Path planning was executed according to Alg.\ \ref{PathPlanning}, as described in section \ref{PathPlanningImp}. The following parameter values were used in all experiments: 
$\backwardRate=0.1$, $\backwardDecay=10^3$, $\lambda=10^3$, and max\_steps $= 80$.


We performed several experiments to evaluate the role of different parameters in our method. Specifically, we independently vary the dimension of the reduced space $\assocSpace^\prime$, the width of bundles $\phi$, the size of the training set, the resolution of the neural map, and the way that bundles are formed. In each case other parameters are held fixed.


In some experiments, we used a single fixed start point, with an outstretched arm, for all trajectories (Fig.\ \ref{fig:arc}). This restriction reduced the period of motor babbling and possible damage to the robot for these experiments. When a fixed starting point was used, $700$ random trajectories were generated on the arc in front of the robot for training. For testing, $300$ goal points were generated on the same arc. 

In other experiments, we used multiple fixed start points, with the outstretched arm in eight different starting points (Fig.\ \ref{fig:arc}). This flexibility provided a more robust learning environment. For each of the eight start positions, we generated $300$ training trajectories stopping at random points of the same arc, for a total $2400$ trajectories. (This number was based on results from the single fixed starting point trials, as mentioned below.) For testing, we generated $150$ goal points on the arc for each of the eight start positions, for a total $1200$ points. 


The fixed starting position method was used for the experiments varying the dimension of $\assocSpace^\prime$ and the size of the training set, and the variable starting position method was used for the experiments varying the resolution of the neural map and the method of bundle formation. The experiment varying the width of bundles $\phi$ was performed according to both methods.


\begin{figure*}[!t]
  \centering
 \subfloat[]
 {\includegraphics[scale=0.45]{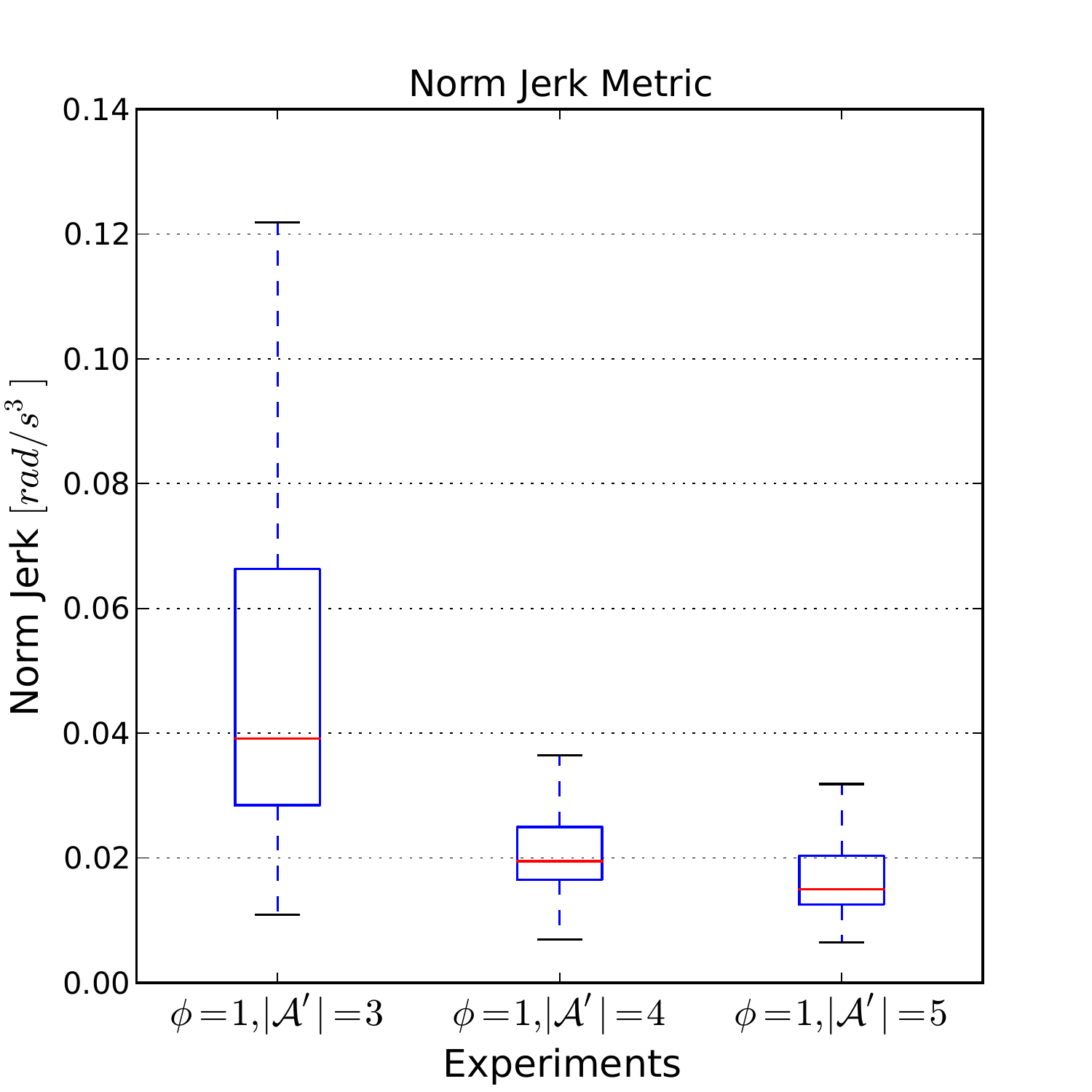}  \label{fig:norm-jerk}} 
 \subfloat[]
 {\includegraphics[scale=0.45]{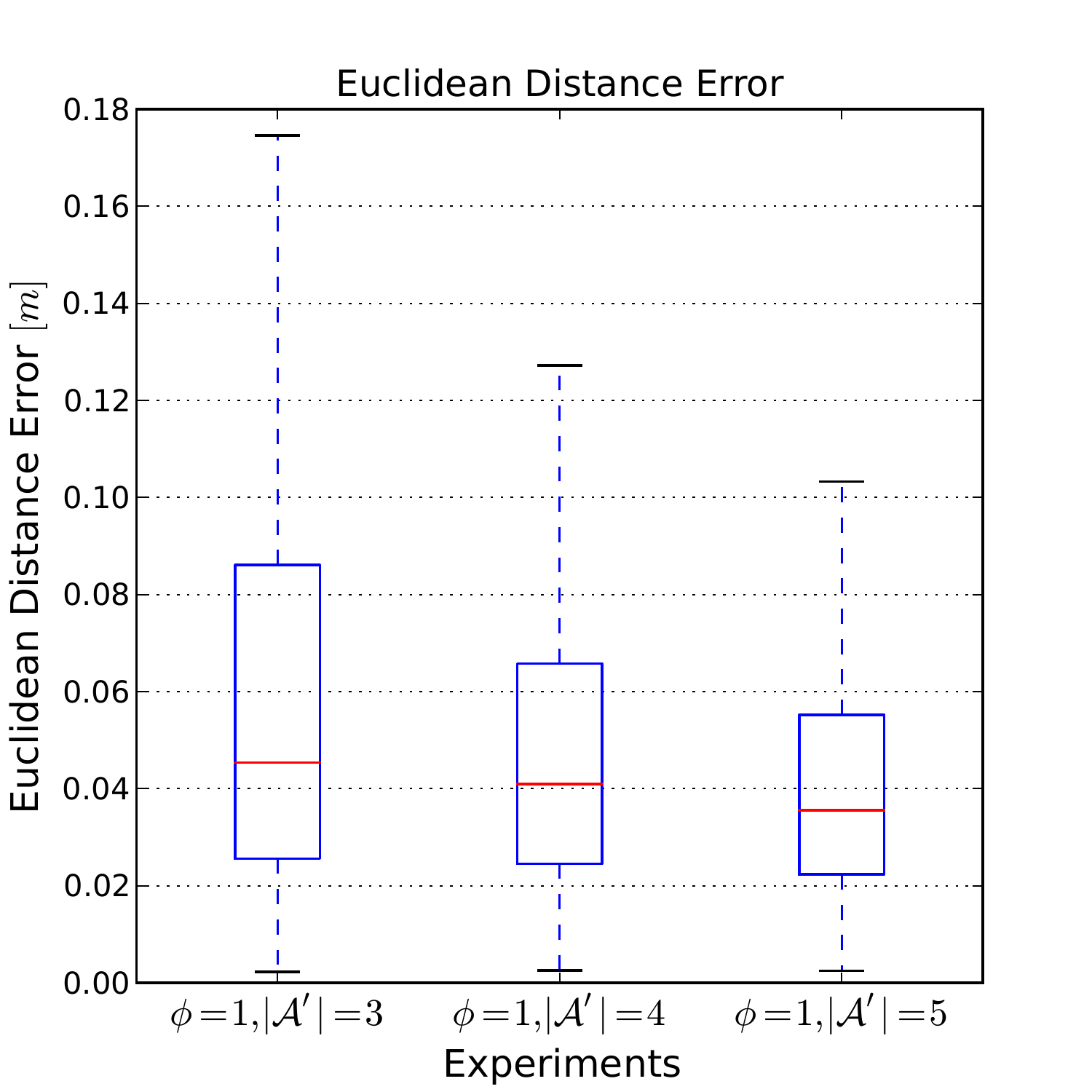} \label{fig:euc-dist}}\quad
  \subfloat[]
 {\includegraphics[scale=0.45]{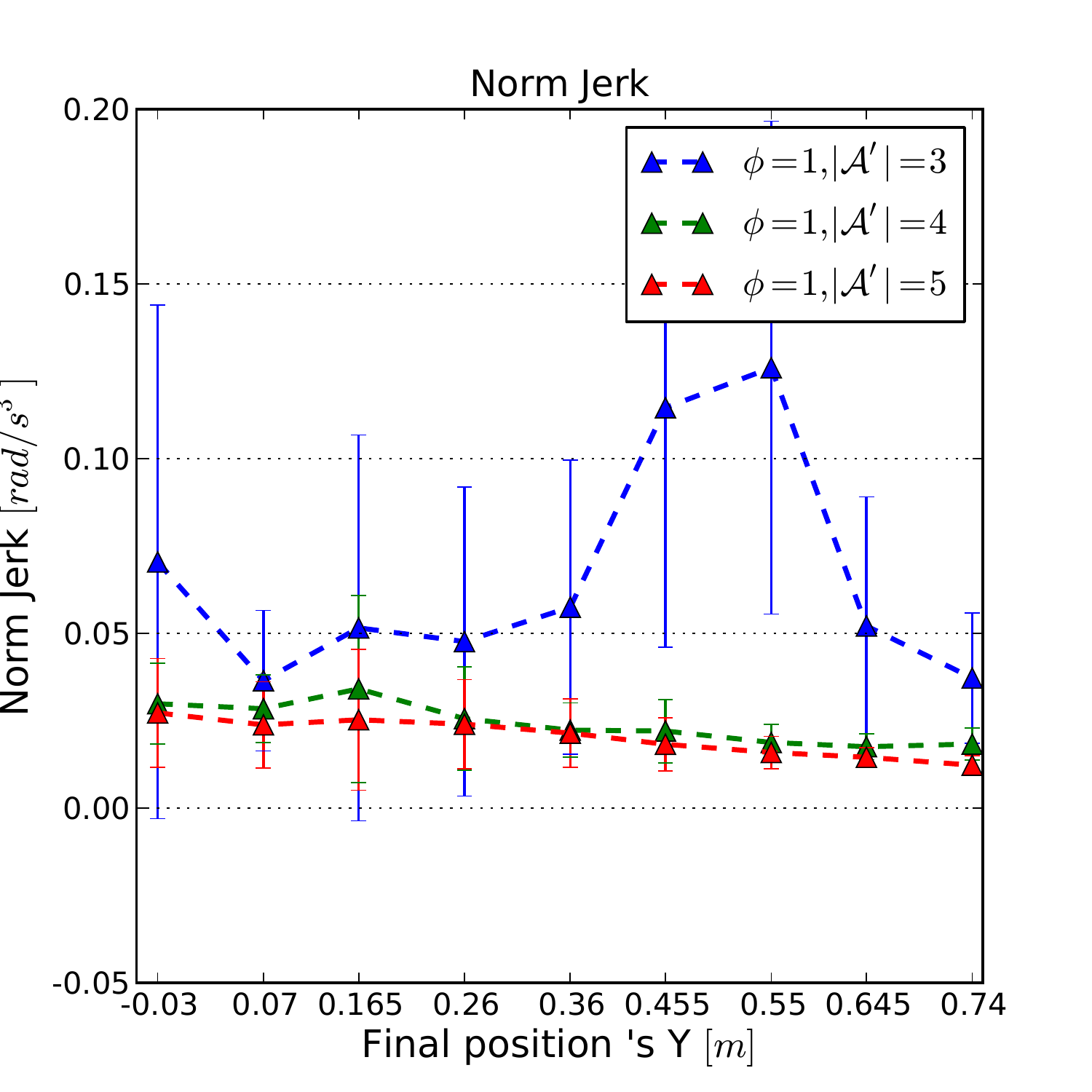}  \label{fig:mean-norm-jerk}}
  \subfloat[]
 {\includegraphics[scale=0.45]{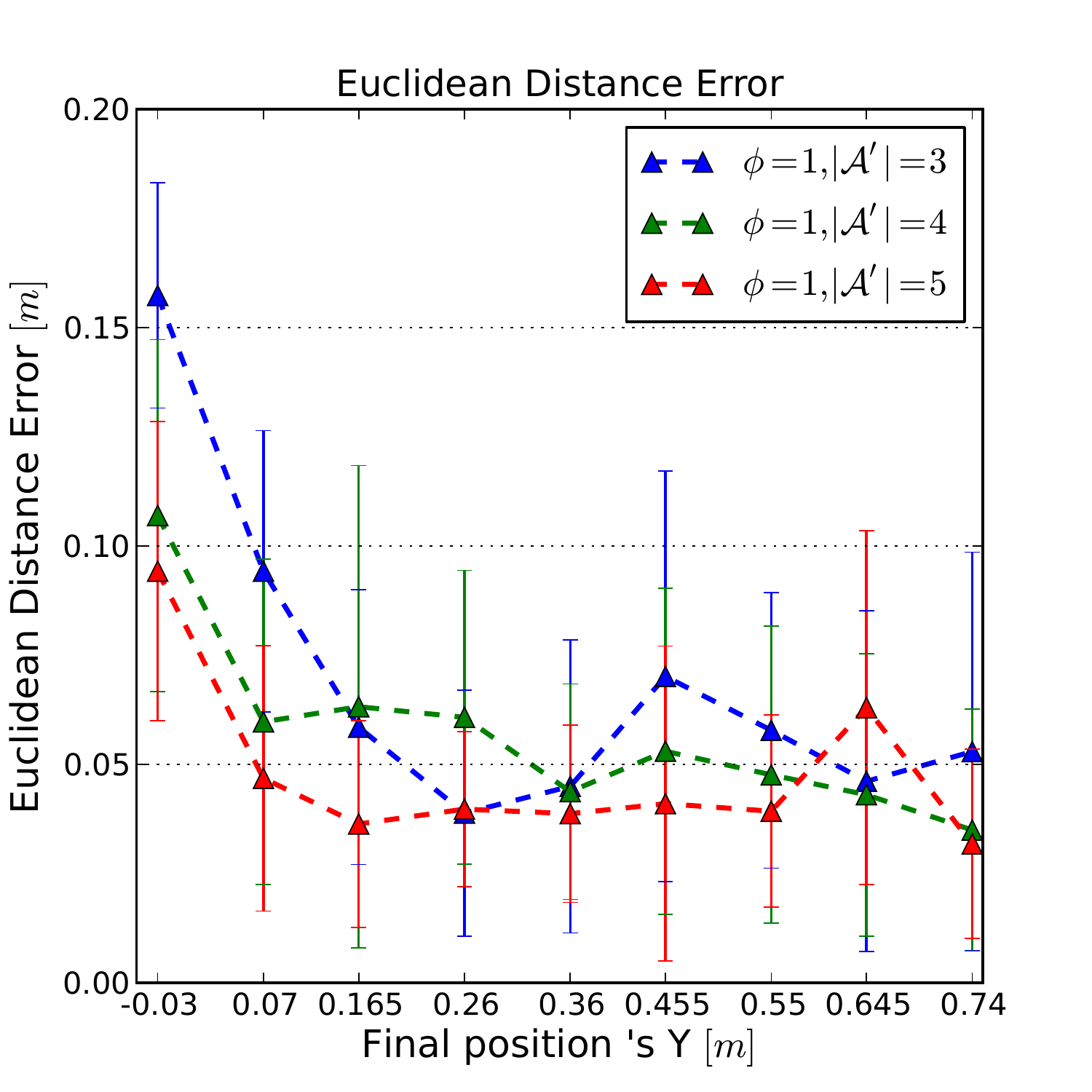}  \label{fig:mean-euc-dist}}
 \caption{Evaluation of planned motions for three trials, where $|\phi|=1$, and training is performed with a single fixed start: (1) $|\assocSpace^\prime|=3 $, (2) $|\assocSpace^\prime|=4 $, and (3) $|\assocSpace^\prime|=5$. (a) Smoothness of planned trajectories based on norm jerk metric across three trials. (b) Euclidean distance error of end effector position across three trials. (c) Smoothness of planned trajectories based on norm jerk metric across $y$ component of final positions in the robot's coordinate system. (d) Mean Euclidean distance error of end effector position across $y$ component of final positions.}
\end{figure*}

For each experiment, we evaluated the planned arm motions during the test with two metrics:
\begin{itemize}
\item {\it End effector distance} estimates the accuracy of the reaching test as the distance in 3D space between the target position and the end effector after reaching is complete, that is, $||g_{(x,y,z)}-\tilde{g}_{(x,y,z)}||$, where $g_{(x,y,z)}$ is the desired location of the end effector and $\tilde{g}_{(x,y,z)}$ is the resulting location. 
\item {\it Norm jerk} evaluates the smoothness of the reaching trajectories in the joint space based on the time derivative of the joint angle acceleration; it is defined  
${\rm jerk} =\frac{1}{f} \sum_{t=1}^{f} ||\dddot \MSpoint_t||$. 
In this equation, $\MSpoint_t \in \assocSpace$ is a point along a planned trajectory and $f$ is the number of points in the trajectory.
\end{itemize}


\subsection{Varying dimension of the reduced space}
In this experiment, we varied the size of the reduced space $\assocSpace^{\prime}$ from three to five dimensions (\textit{i.e.}, three, four or five neurons in the bottleneck of the autoencoder). In the second phase of training, neural maps were built for each of these dimensionalities of space $\assocSpace^\prime$. In all trials, the width parameter $\phi$ was set to $1$ to create narrow bundles. 

Fig.\ \ref{fig:norm-jerk} illustrates the norm jerk of planned trajectories for different dimensionalities of space $\assocSpace^\prime$. As shown in Fig.\ \ref{fig:norm-jerk}, the median of norm jerk decreases as the size of space $\assocSpace^\prime$ increases. This metric suggests that the smoothness of planned motions depends on the accuracy of the autoencoder. However, the range of norm jerk shows a nonlinear drop from size $3$ to sizes $4$ and $5$. This nonlinear drop in the norm jerk suggests that increasing the size of space $\assocSpace^\prime$ might not change the smoothness of motions beyond a certain size. Fig.\ \ref{fig:euc-dist} shows error in end effector position across the three dimensionalities of space $\assocSpace^\prime$. The error linearly decreases as the dimensionality of space $\assocSpace^\prime$ increases. Together, Figs.\ \ref{fig:norm-jerk} and \ref{fig:euc-dist} show that the model has successfully learned to plan motions from the test set accurately and smoothly when the size of space $\assocSpace^\prime$ is $4$ or $5$.

Fig.\ \ref{fig:mean-norm-jerk} shows the mean and standard deviation of norm jerk across the $y$ component of final positions in the robot's coordinate system (across the torso). The motions are smoother across the $y$ component for a space $\assocSpace^\prime$ of size $4$ or $5$. Fig.\ \ref{fig:mean-euc-dist} displays the mean and standard deviation of the Euclidean distance of end effector positions from goal positions across the $y$ component of final positions. This metric also indicates that a space $\assocSpace^\prime$ of size $4$ or $5$ tends to produce more accurate planned motions; however, there is no particular trend in the error as $y$ increases. An interesting anomaly in this figure is that the error is higher for small $y$ values, that is, short trajectories. On investigation, we found that a set of movements to the right side of the arc involves positive values of the second joint of the shoulder. It seems that these border values were not learned accurately in the autoencoder, and the model is not able to represent the poses that are located to the right of the arc as well as it can those on the left side.

\subsection{Varying training set size}
\begin{figure}[!t]
\centering
\includegraphics[scale=0.45]{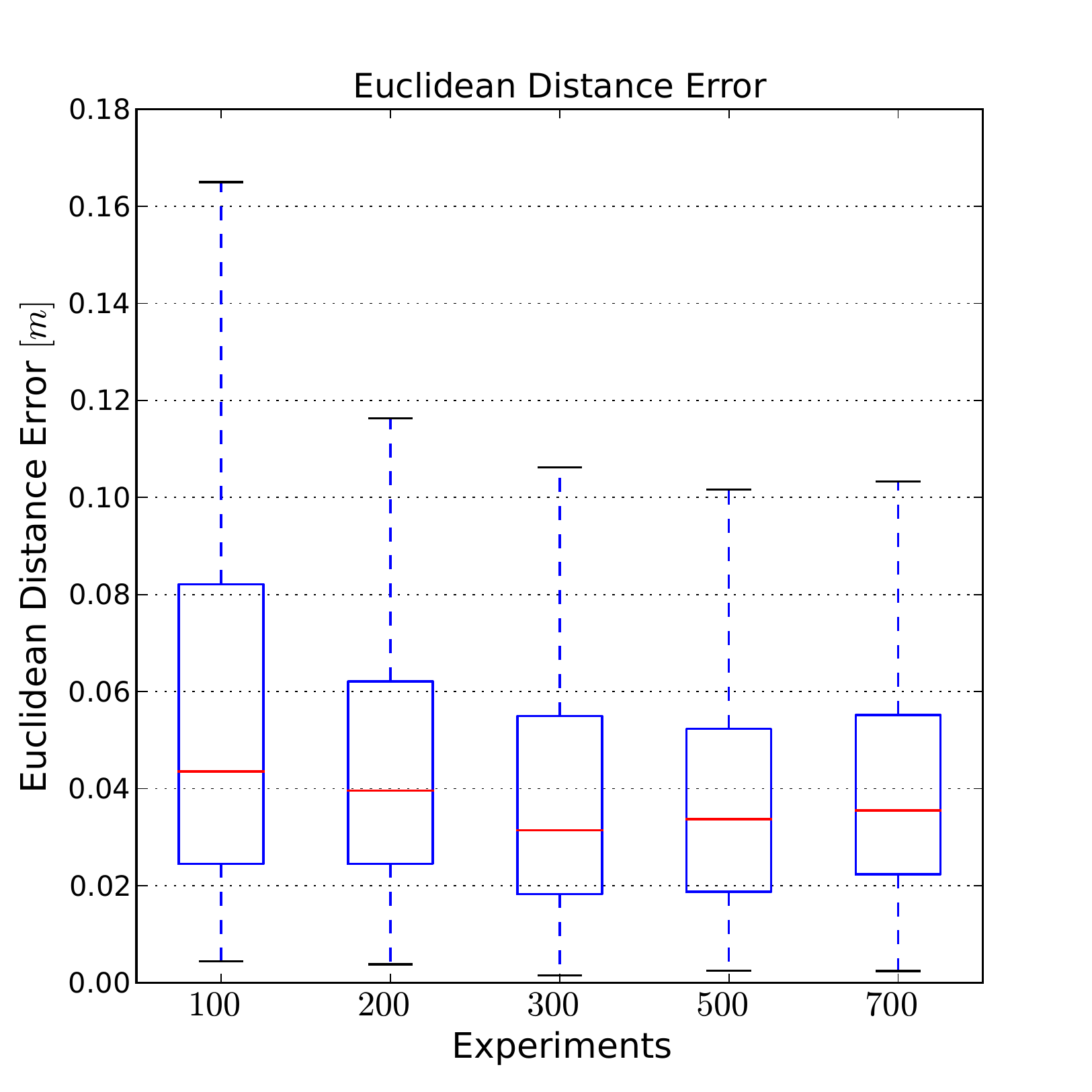}
\caption{Evaluation of planned motions for five trials, where $|\phi|=1$, $|\assocSpace^\prime|=5$, and training is performed with a single fixed start: (1) $train=100 $, (2) $train=200 $, (3) $train=300 $, (4) $train=500 $, and (5) $train=700 $. Euclidean distance error of end effector position across five trials}
\label{fig:varying-train-size}
\end{figure}

To find the number of babbling trajectories needed for the system to master this simple task, we trained the autoencoder with different training sizes. Specifically, $100$, $200$, $300$, $500$, and $700$ training trajectories were used in five trials. In each trial the testing size was fixed at $300$ goal points, the bundle width was $1$, and the dimensionality of the reduced space $\assocSpace^\prime$ was $5$. Table \ref{tab:autoencoder} shows that as the size of the training set increases the accuracy of the autoencoder improves both regarding variance and root-mean-square error of the test set. In all these trials, the bottleneck of autoencoder was set to $5$. However, the accuracy of the autoencoder doesn't change significantly after the training set of $300$ babbling trajectories. 
Fig.\ \ref{fig:varying-train-size} shows the end effector position error for the different training sizes. As training size increases, error diminishes until the training size of $300$ where the accuracy doesn't notably improve. We used this portion as a basis of the next set of experiments with multiple fixed starting points. 
\begin {table}[!t]
\caption{Accuracy of Autoencoders with Different Training Size.}
\label{tab:autoencoder}
\begin{center}
 \begin{tabular}{|c c c|} 
 \hline
 Train set size & Test RMSE  & Test Variance\\ [0.5ex] 
 \hline\hline
 100 & 0.21 &  0.91\\ 
 \hline
 200 & 0.21 & 0.91\\
  \hline
 300 & 0.04 &  0.96\\
 \hline
  500 & 0.049 &  0.97\\
 \hline
 700  & 0.049 &  0.97\\
 \hline
 \end{tabular}
\end{center}
\end{table}

\subsection{Varying bundle width}
\begin{figure*}[!t]
  \centering
 \subfloat[]
 {\includegraphics[scale=0.45]{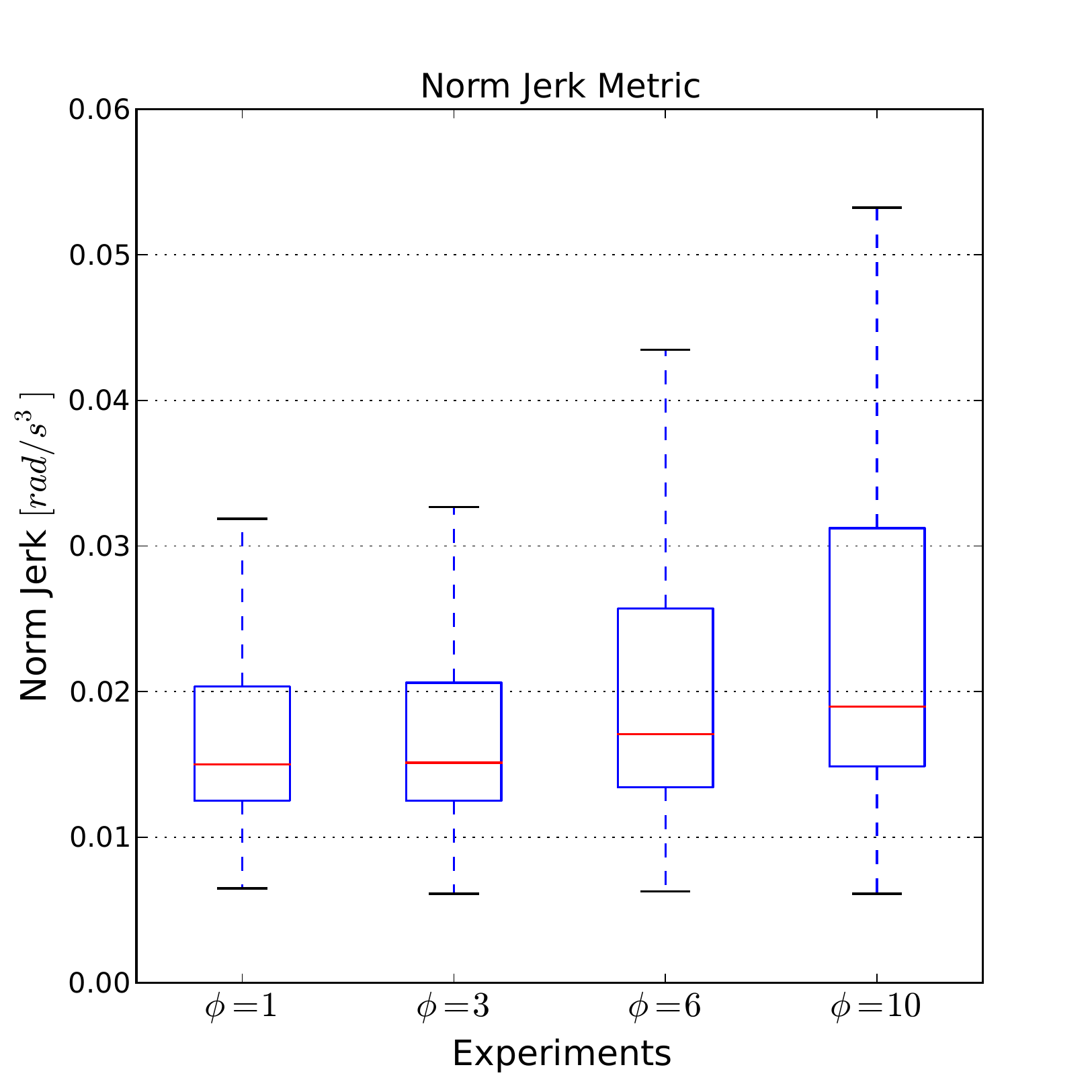}  \label{fig:norm-jerk-phi}} 
 \subfloat[]
 {\includegraphics[scale=0.45]{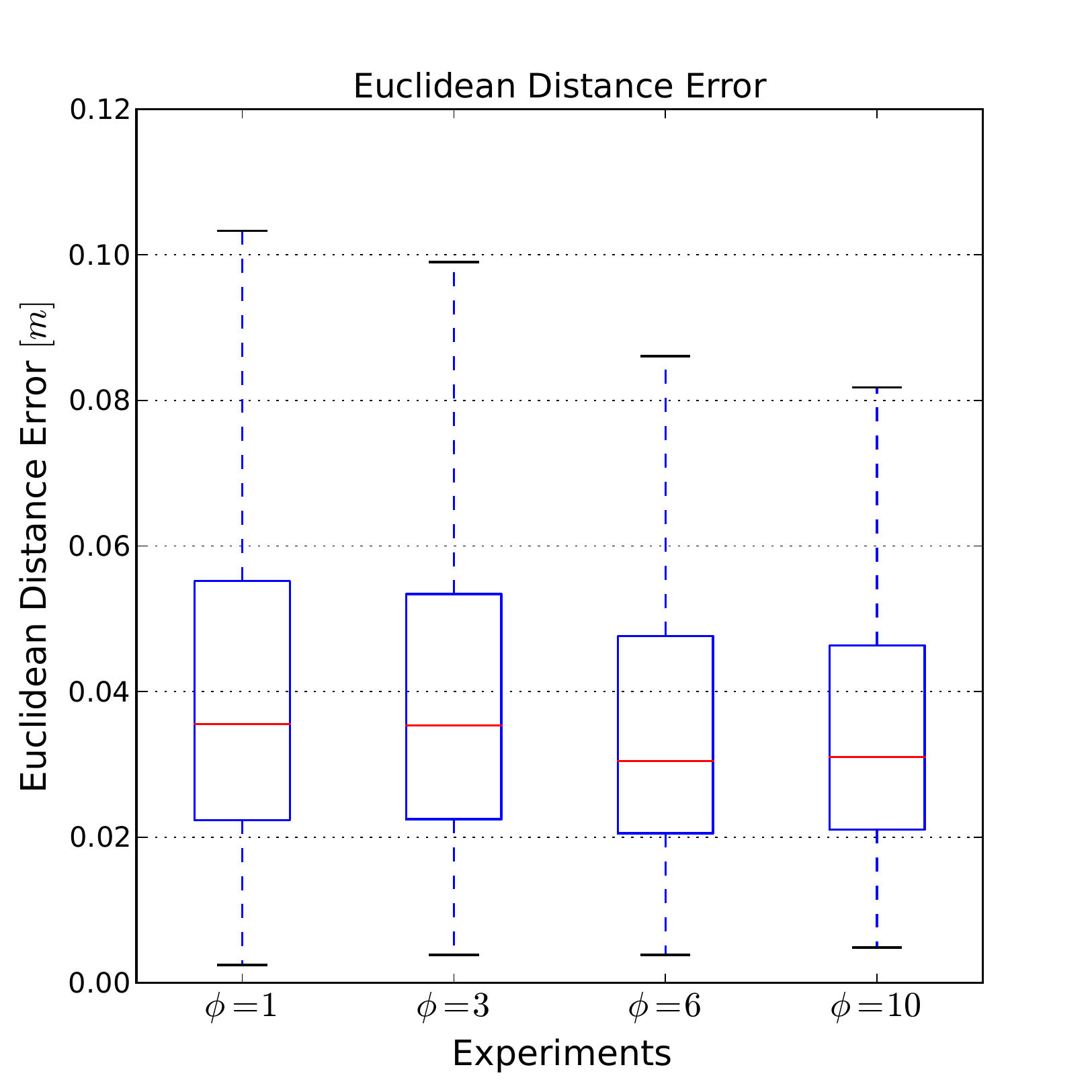} \label{fig:euc-dist-phi}}\quad
  \subfloat[]
 {\includegraphics[scale=0.45]{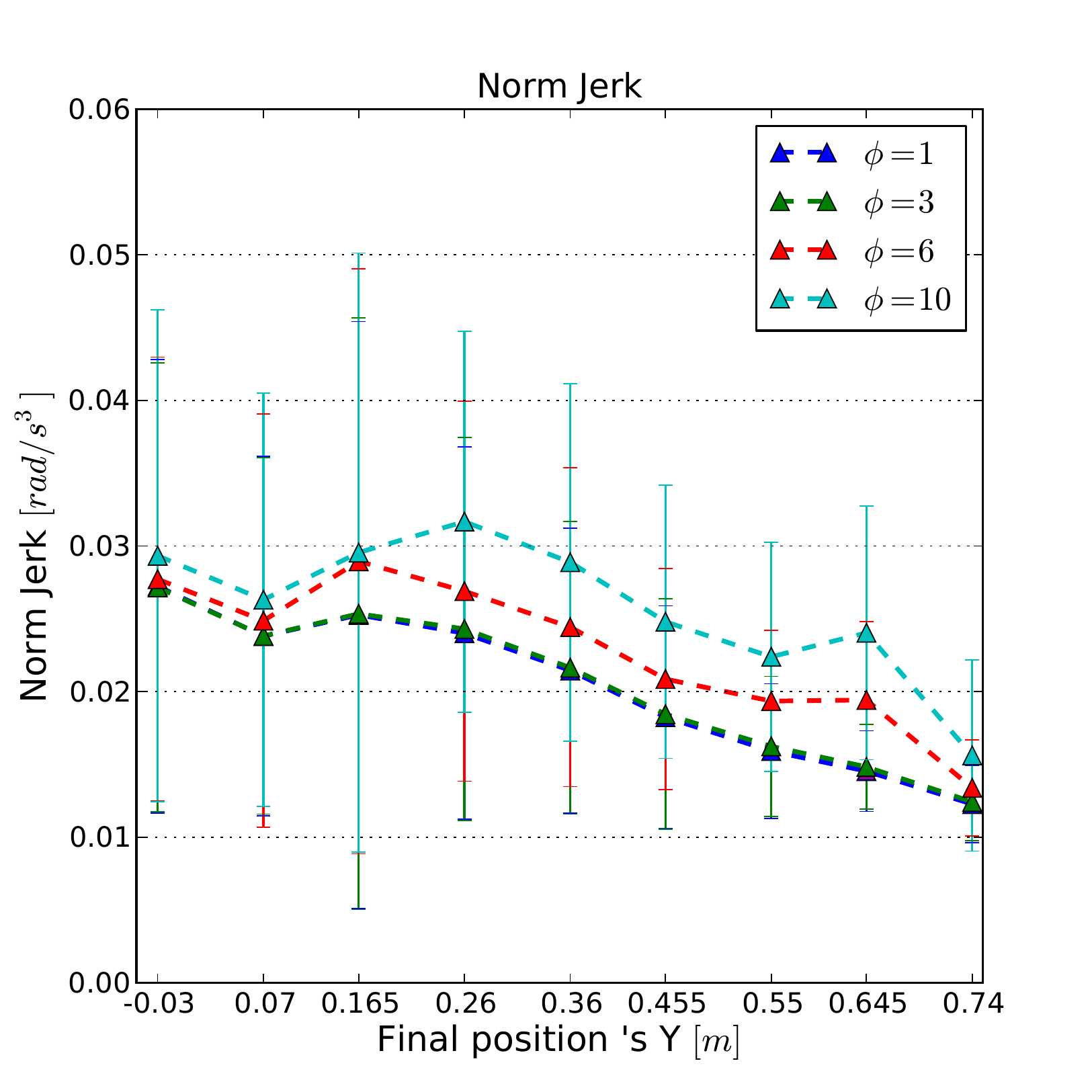}  \label{fig:mean-norm-jerk-phi}}
  \subfloat[]
 {\includegraphics[scale=0.45]{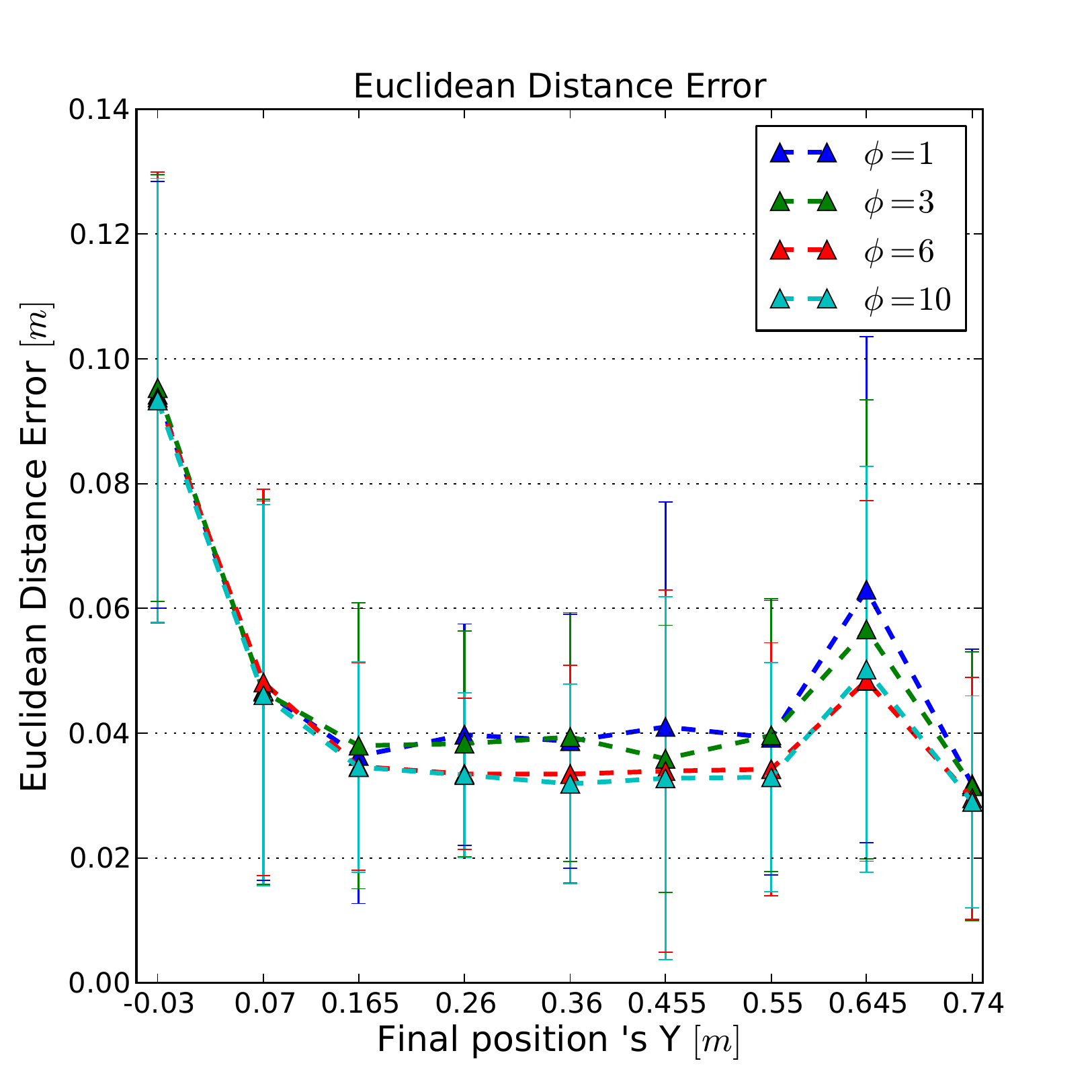}  \label{fig:mean-euc-dist-phi}}
 \caption{Evaluation of planned motions for four trials, where $|\assocSpace^\prime|=5$, and training is performed with a single fixed start: (1) $|\phi|=1 $, (2) $|\phi|=3 $, (3) $|\phi|=6$, and $|\phi|=10$. (a) Smoothness of planned trajectories based on norm jerk metric across four trials. (b) Euclidean distance error of end effector position across four trials. (c) Smoothness of planned trajectories based on norm jerk metric across $y$ component of final positions in the robot's coordinate system. (d) Mean Euclidean distance error of end effector position across $y$ component of final positions.}
\end{figure*}

In two experiments, we tested the effect of different bundle widths $\phi$, once with single and once with multiple starting positions.
For the single starting position, we tested four different bundle widths $\phi$: $1$, $3$, $6$, and $10$. Dimensionality of the reduced space $\assocSpace^\prime$ was $5$. Fig.\ \ref{fig:norm-jerk-phi} shows the norm jerk of planned trajectories for these trials. As bundles become wider, the median of norm jerk increases slightly. Fig.\ \ref{fig:euc-dist-phi} shows end effector position error, which decreases slightly as bundles becomes wider. The position error doesn't change much as $\phi$ changes from $1$ to $3$, nor from $6$ to $10$; instead, for both metrics the important change occurs between $\phi$ of $3$ and $6$.
This shows that the model was able to generalize better with wider bundles, but the broader bundle also has had a negative impact on the smoothness of trajectories. 

Fig.\ \ref{fig:mean-norm-jerk-phi} shows the mean and standard deviation of norm jerk across the $y$ component of final positions in the robot's coordinate system. The smoothness of trajectories improves consistently as $y$ increases (that is, longer motions), and improves as bundle width decreases down to $\phi=3$. The fact that the shorter trajectories are less smooth than the longer ones suggests that the autoencoder was not able to learn some of the joint positions that are more prevalent on the right side of the arc. Fig.\ \ref{fig:mean-euc-dist-phi} shows the mean and standard deviation of the end effector position error from goal positions across the $y$ component. This metric also indicates that bundle widths $\phi$ of $6$ or $10$ tend to produce more accurate planned motions across the $y$ values except for the movements that lead to the right side of the arc ($y < 0.07$).
These shorter trajectories are in the boundary, and generalizing the babbling trajectories will not help to find a better motion on the very far right side of the babbling arc. 

\begin{figure*}[!t]
  \centering
 \subfloat[]
 {\includegraphics[scale=0.45]{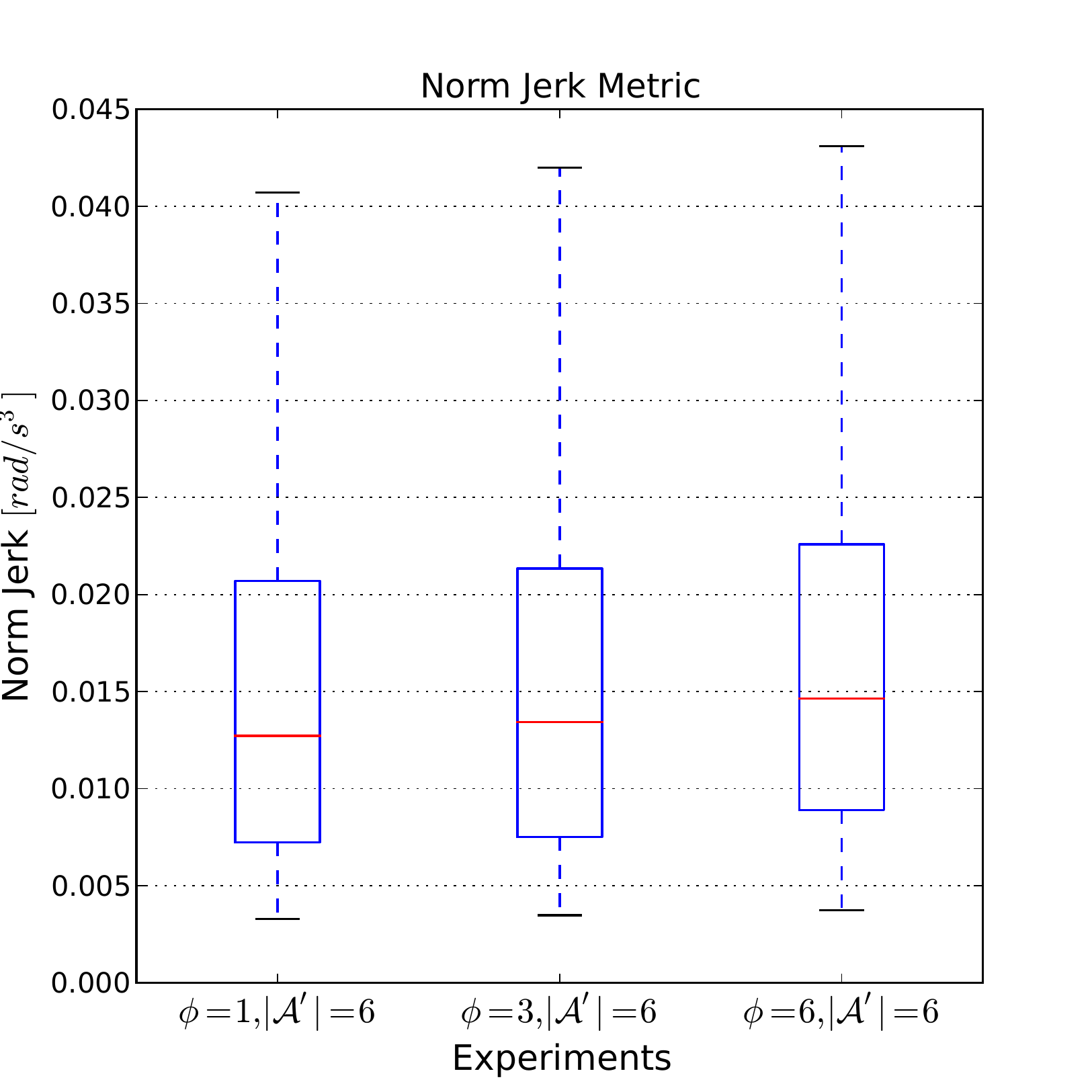}  \label{fig:norm-jerk-multi-starts}} 
 \subfloat[]
 {\includegraphics[scale=0.45]{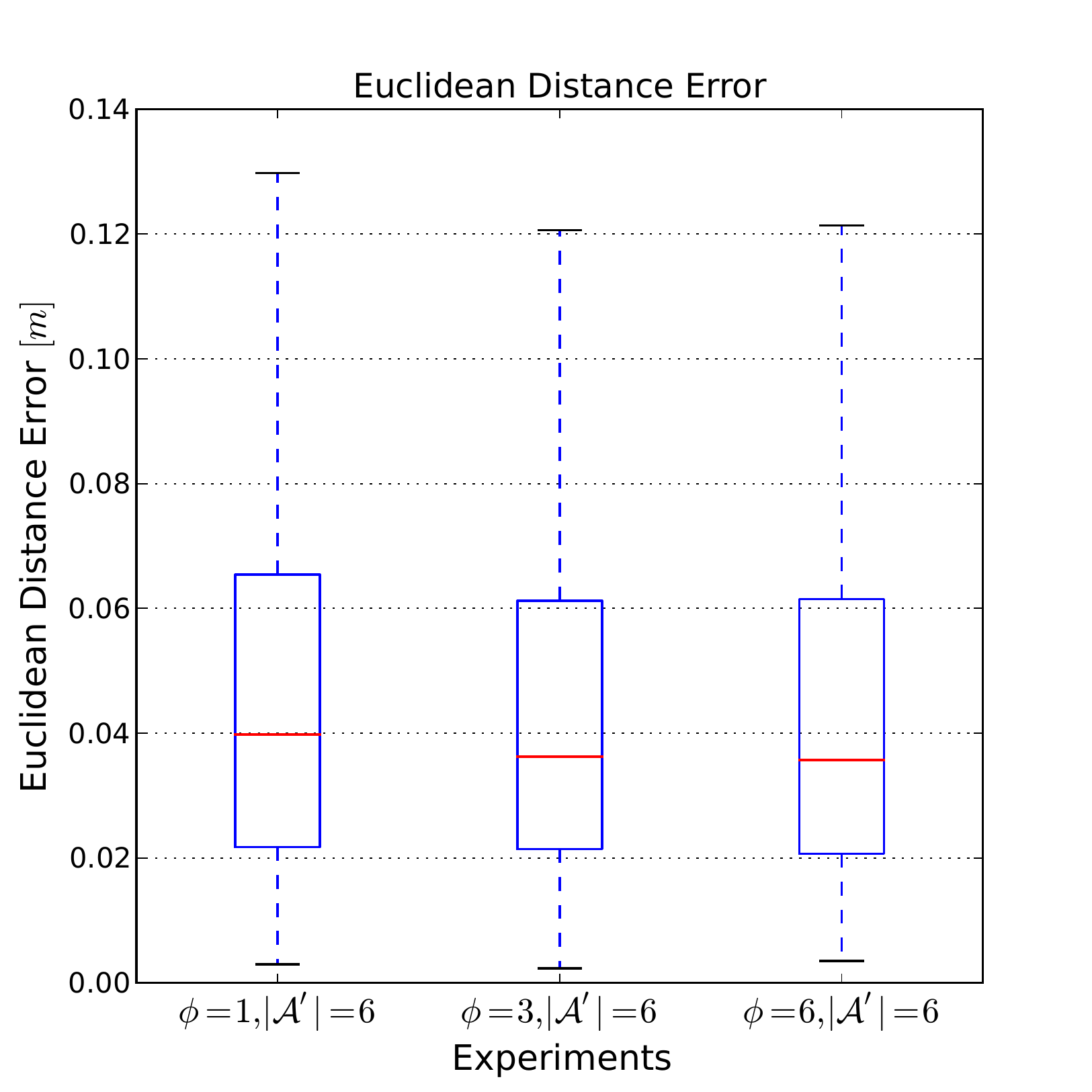} \label{fig:euc-dist-multi-starts}}\quad
 \caption{Evaluation of planned motions for three trials, where $|\assocSpace^\prime|=6$ is set for the trials and training is performed for various fixed start points: (1) $|\phi|=1 $, (2) $|\phi|=3 $,and (3) $|\phi|=6$. (a) Smoothness of planned trajectories based on norm jerk metric across three trials. (b) Euclidean distance error of end effector position across three trials. }
\end{figure*}

For variable starting positions, we tested the three bundle widths $1$, $3$, and $6$, this time with a space $\assocSpace^\prime$ of size $6$. The remaining parameters of the training and path planning are the same as the experiment with a single fixed start. Fig.\ \ref{fig:norm-jerk-multi-starts} shows the norm jerk of resulting planned trajectories. As the bundle width increases, median norm jerk increases. Fig.\ \ref{fig:euc-dist-multi-starts} shows that end effector position error decreases as the bundle width increases from $1$ to $3$ or $6$. As was the case when start position was fixed, these graphs show that the model generalizes better with wider bundles, but the broader bundles also have a negative impact on the overall smoothness of trajectories.

\subsection{Varying resolution of the neural map}
To investigate how the neural map's structure might affect the ability of the algorithm to generalize, we tested the effect of the neural map's resolution on the resulting planned trajectories.
We tested three resolutions: the median difference between features of the consecutive points in trajectories, which we call $\rm medRes$, and then this resolution multiplied by two and by three ($2\cross \rm medRes$ and $3\cross \rm medRes$, respectively). (By ``resolution'' we mean difference between values represented by adjacent neurons in the map; thus, $3\cross \rm medRes$ is the most coarse resolution.)
In each trial, bundle width $\phi$ was set to $3$ and size of space $\assocSpace^\prime$ to $6$.

\begin{figure*}[!t]
  \centering
 \subfloat[]
 {\includegraphics[scale=0.46]{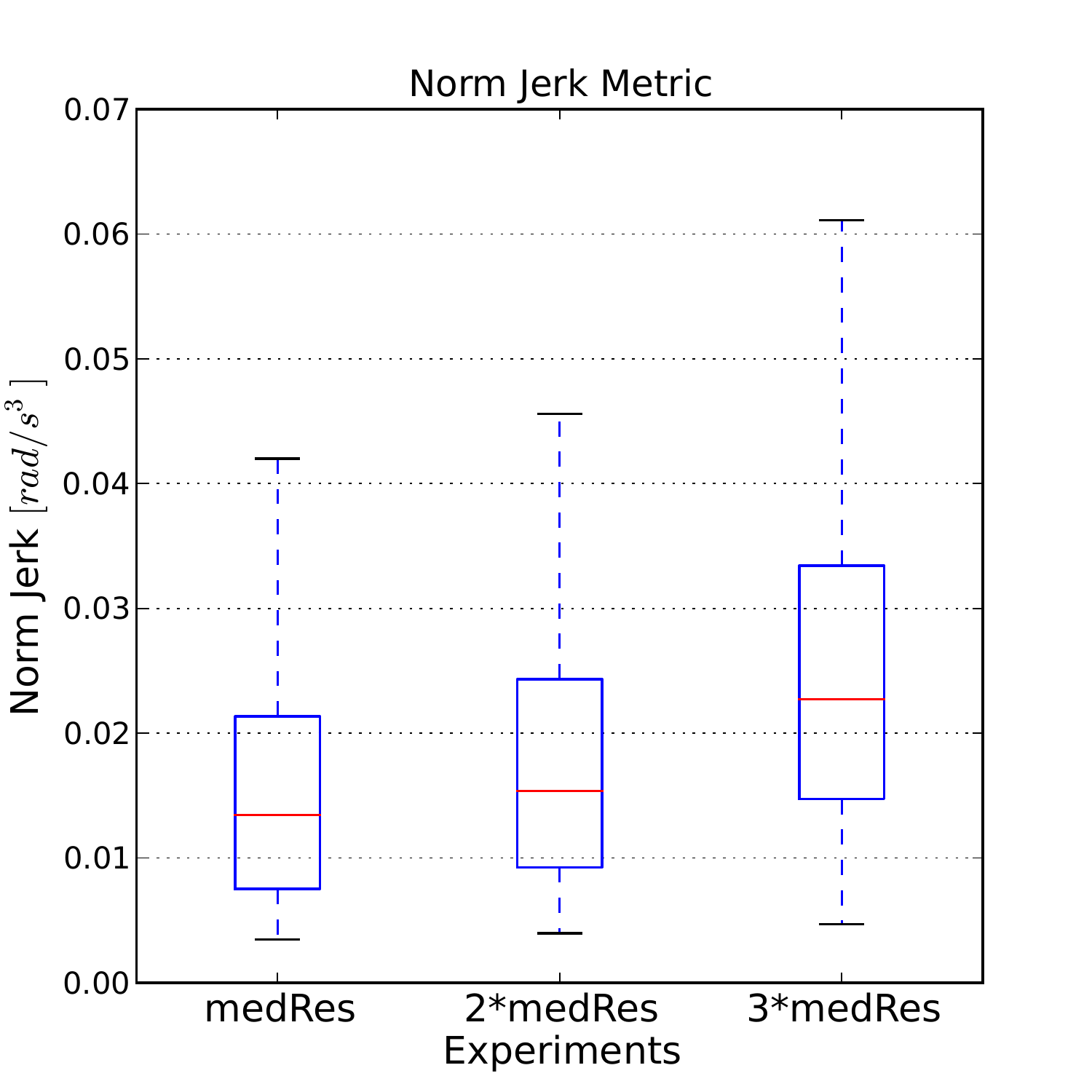}  \label{fig:jerk-modifyRes}} 
 \subfloat[]
 {\includegraphics[scale=0.46]{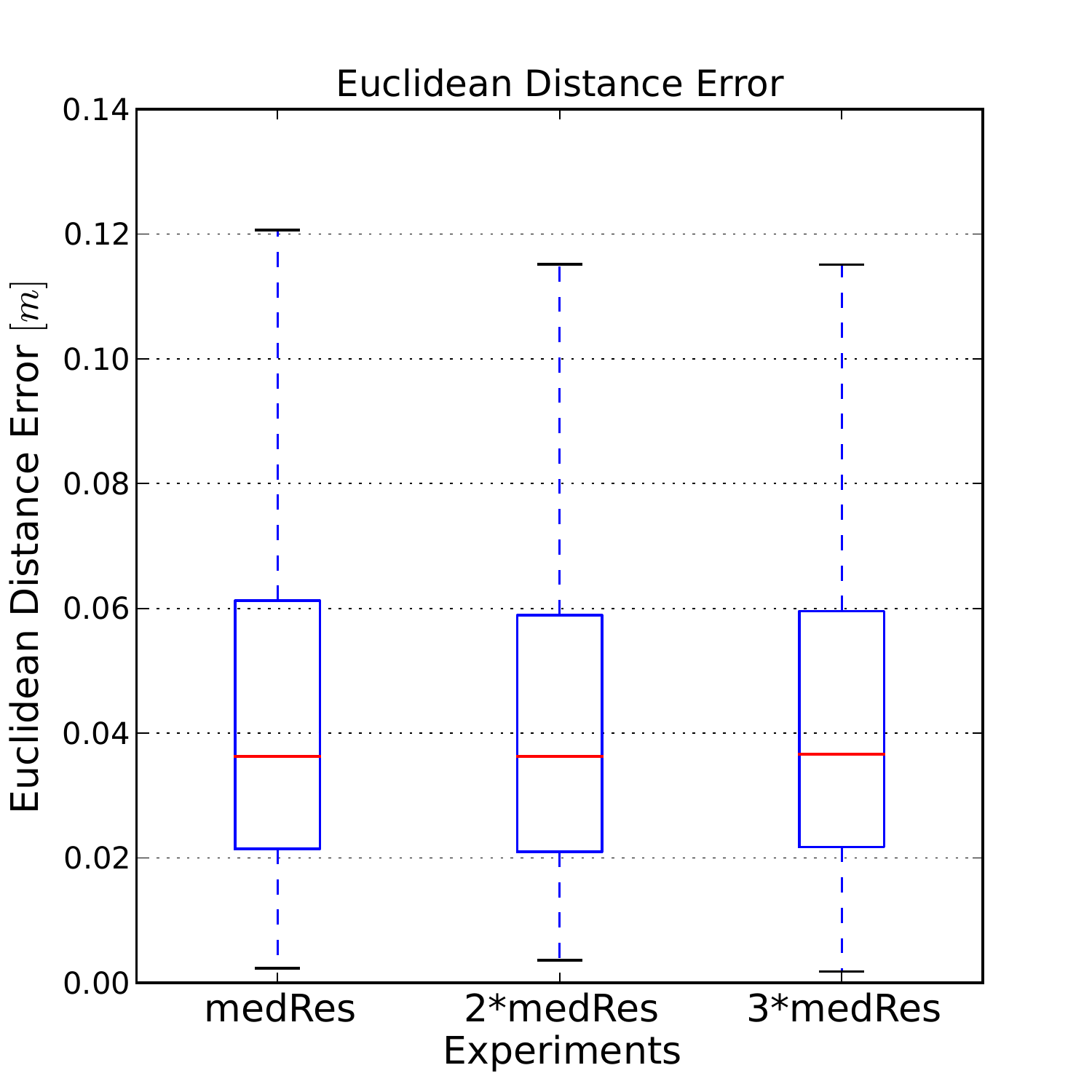} \label{fig:euc-modifyRes}}\quad
 \caption{Evaluation of planned motions for three trials, where $|\phi|=3$, $|\assocSpace^\prime|=6$, and training is performed for various fixed start points: (1) the neural map's resolution was median of difference between features (2) the neural map's resolution was $2 \cross$ the resolution of the first trial. (3) the neural map's resolution was $3 \cross$ the resolution of the first trial in terms of (a) Smoothness of planned trajectories based on norm jerk metric across three trials. (b) Euclidean distance error of end effector position of planned trajectories across three trials.}
\end{figure*}

Fig.\ \ref{fig:jerk-modifyRes} shows the norm jerk of planned trajectories for the three neural map resolutions. Trajectories become less smooth with increasing coarseness of map resolution.
Fig.\ \ref{fig:euc-modifyRes} shows end effector position error for the three trials. There is almost no difference in position error with changing resolution.
Based on the jerk metric, it seems that in a map with coarse resolution, neurons are more general (respond to a larger region of the motor-sensory space) and cannot represent finer details needed for a smoother path. However, the end effector position error doesn't reflect this finding.

\subsection{Varying bundle formation}
\label{sec:varBundleForm}
In this experiment, we compared two modifications of the bundle formation that we call $\rm fixConnections$ and $\rm parConnections$ with the proposed one in Alg.\ \ref{BundleLearning} ($\rm lnrConnections$) in terms of smoothness. In these trials, $\phi=3$ and $|\assocSpace^\prime|=6$. For $\rm lnrConnections$, which we consider our baseline for these trials, bundles of width $3$ are formed with central connections weighted more than the synapses to the side. We reduced the weights linearly from the center to the sides. For $\rm fixConnections$, bundles of width of $3$ are formed, but the connections are uniform and have the fixed weight of $1.0$ throughout the bundles. For $\rm parConnections$, bundles of width $3$ are formed in a new fashion. In our usual algorithm, we set up connections among all the neurons that are activated at time $i$ and all the neurons that are fired at time $i+1$ (Alg.\ \ref{BundleLearning}). In this third trial, we ranked the firing neurons at each time step based on their firing rates. The connections from step $i$ to $i+1$ are only formed among the neurons with the same rank. As an analogy, this approach tends to create parallel streams within a bundle. 

\begin{figure}
\centering
\includegraphics[scale=0.45]{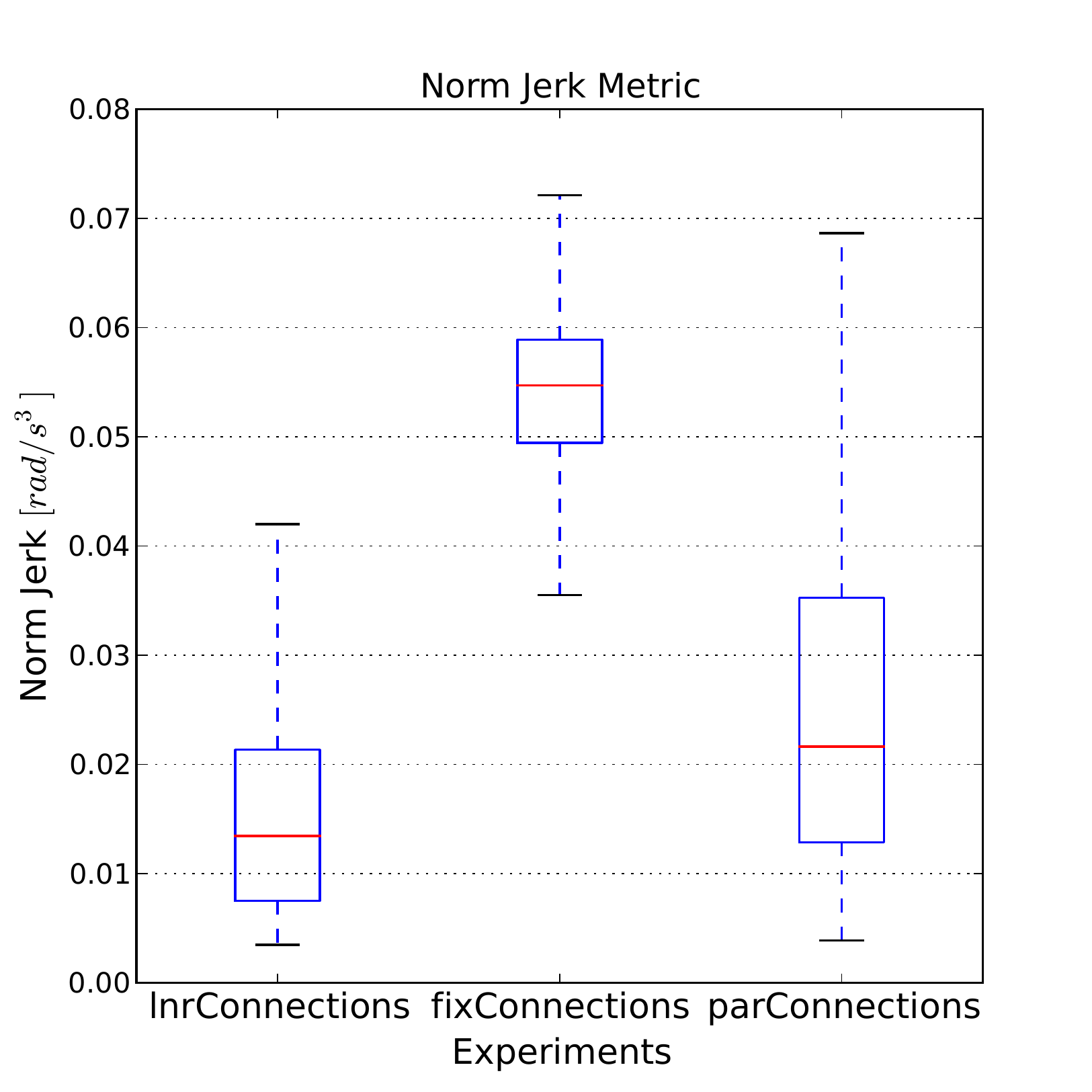}
\caption{Evaluation of planned motions for three trials, where $|\phi|=3$, $|\assocSpace^\prime|=6$, and training is performed for various fixed start points: (1) linearly decreasing connections in bundles, (2) fix-strength connections in bundles, (3) parallel connections in bundles. Smoothness of planned trajectories based on norm jerk metric across three trials. }
\label{fig:modify-bundles}
\end{figure}

End effector position error was the same across these trials, so we only present results related to smoothness.
Fig.\ \ref{fig:modify-bundles} shows norm jerk of planned trajectories across the three trials. $\rm fixConnections$ produced the least smooth motions compared to the others, which might suggest that non-uniform bundles has been more effective in path planning in this model. $\rm parConnections$ also produced more jerky motion compared to the baseline $\rm lnrConnections$. The average of norm jerk is $0.015$ for the baseline $\rm lnrConnections$ and $0.02$ for $\rm parConnections$.
The box-and-whisker plots for these two trials illustrate that $\rm parConnections$ is less favorable for some of the test trajectories but not all. This finding may also suggest that denser bundles with more neural connectivity help path planning in our model a better chance of finding an optimal path.

\section{DISCUSSION}
\label{discussion}
In this paper, we analyzed aspects of our model including dimension reduction, bundle formation, the amount of motor exploration needed, and the resolution of neural maps. 

One important facet of our model is the design and resolution of the neural map. We believe there is a trade-off between the resolution of a neural map and the required time for motor exploration to find a path. A neural map with fine resolution requires more neurons to span the motor-sensory space while a map with coarse resolution demands fewer neurons. In a map with coarse resolution, neurons cannot represent finer details of reaching because multiple motor-sensory points are assigned to one neuron. These neurons respond to a larger area of the motor-sensory space; thus, the planned motions can become less accurate. On the other hand, a map with finer resolution can represent more details of motor-sensory space and can produce more accurate movements, but demands more exploration effort.

Another important factor in our model is the trade-off between the required amount of babbling and the required amount of generalization in the neural map. It is clear that more motor babbling allows an agent to explore the motor-sensory space more thoroughly, but what is the right amount of babbling, and how much of the space is searched by infants through motor exploration? We believe that it is not possible for a realistic agent to explore the entire motor-sensory space, which can be vast. Therefore, to achieve the same performance, either babbling should be increased or the learning mechanism must be more general. In our model, increasing bundle width is an attempt to generalize without the need to explore the entire motor-sensory space. We are not entirely sure how infants unconsciously generalize from an experienced motion to another feasible motion with no loss of accuracy. In future work, we hope to find bundle formation methods that more fruitfully generalize from limited babbling data to other possible trajectories without sacrificing accuracy. 

We are focused on a single stage of development, the encoding of information resulting from motor babbling into sparse neural maps representing motor-sensory space. One reason we are interested in this stage is because it is embodied and emergent, representing an interplay between a genetic program and its environment.
Information about the environment becomes encoded implicitly as information about body-environment interactions. 
Other important (especially earlier) stages of development seem more genetically determined, such as forming the topology of the motor-sensory maps. We treat this topology as a given in our model. In the future we hope to integrate the present model into a broader, multi-stage developmental model.
One step in this direction would be to impose more structure on the motor-babbling stage. 

Our model forms bundles of connected neurons in which synapse weights are strongest in the middle and decrease linearly in strength toward the edges. In section \ref{sec:varBundleForm}, we tested two modifications to this approach. The first was to form connections of uniform strength throughout the bundle. The second was to create less dense bundles as explained in section \ref{sec:varBundleForm}. Each of these modifications resulted in less smooth planned motions. These results suggest that more links within in a bundle with non-uniform connections give the diffusion in path planning a better chance of finding an optimal path. 

The spreading activation that is responsible for motion planning in our model requires a careful choice of activation rate and decay rate parameters. An activation rate that is too high causes activity in the neural map to saturate, meaning no gradient would be present to guide the planning process. Conversely, a decay rate that is too high causes the activation from the goal to fall to zero before reaching the neurons representing the starting motor-sensory state. It would be worthwhile to explore diffusion algorithms that would not require such fine-tuning, for example, a diffusion technique in which the source would not be responsible for producing the gradient. Another disadvantage with this kind of path planning is that the best parameters for long trajectories might not work for shorter ones. So the parameter choice demands investigation with a variety of trajectory lengths. 

In an infant brain, an image of completion as well as a current motor-sensory state would be represented by activity in some group of neurons. For simplicity, our model represents each start and the goal state using a single neuron. In future work we expect to expand our model's representation of motor-sensory states to groups of neurons.

In the real world, real-time autonomous robots expect to learn from new circumstances; therefore, it is important to design an online learning mechanism. However, the current implementation of our model is an offline learning model because the same training trajectories are used in three passes: to find a suitable dimension of motor-sensory space, to create neurons in the neural maps, and to produce synapses among neurons. The choice of using babbling trajectories for creating neurons is not inherent in the model but rather is used in implementation for both convenience and efficiency. In fact, construction of neurons should be offline so the synapses in the neural maps can be modified on the fly during the motor babbling. 

The current model was not tested with any experiments with non-reachable points in the space (e.g. presence of an obstacle). We hope to expand the path planning implementation to account for such situations as follows. In the current implementation, neurons representing a goal are viewed as the attraction forces in the neural map. Similarly, neurons that account for the location of an obstacle could exert inhibitory activities or repulsive forces; therefore at each step, the path planning algorithm should determine motion direction by including those repulsive forces from the obstacle.

In conclusion, we presented an embodied, developmental, and neurally plausible reaching model that is inspired by motor-sensory interaction of an infant and its environment. At the core of this model, three neural maps represent the same motor-sensory space and play different roles in the arm motion planning. The motion planning occurs through the collaboration of these three neural maps, which represent trajectory bundles by means of spreading activation from a goal state. To show that this model is computationally feasible, we tested it in a simple reaching task using a humanoid robot. The model uses motor babbling to acquire smooth and precise reaching behavior. 

\section{ACKNOWLEDGMENT}
The humanoid robot used in this work was acquired through National Science Foundation MRI award 1229176.

\ifCLASSOPTIONcaptionsoff
  \newpage
\fi



\bibliographystyle{./bibtex/IEEEtran}
\bibliography{./bibtex/paper,./bibtex/EFRI,./bibtex/RIbib}

\begin{thebibliography}{10}
\providecommand{\url}[1]{#1}
\csname url@samestyle\endcsname
\providecommand{\newblock}{\relax}
\providecommand{\bibinfo}[2]{#2}
\providecommand{\BIBentrySTDinterwordspacing}{\spaceskip=0pt\relax}
\providecommand{\BIBentryALTinterwordstretchfactor}{4}
\providecommand{\BIBentryALTinterwordspacing}{\spaceskip=\fontdimen2\font plus
\BIBentryALTinterwordstretchfactor\fontdimen3\font minus
  \fontdimen4\font\relax}
\providecommand{\BIBforeignlanguage}[2]{{%
\expandafter\ifx\csname l@#1\endcsname\relax
\typeout{** WARNING: IEEEtran.bst: No hyphenation pattern has been}%
\typeout{** loaded for the language `#1'. Using the pattern for}%
\typeout{** the default language instead.}%
\else
\language=\csname l@#1\endcsname
\fi
#2}}
\providecommand{\BIBdecl}{\relax}
\BIBdecl

\bibitem{mahoor}
Z.~Mahoor, B.~MacLennan, and A.~MacBride, ``Neurally plausible motor babbling
  in robot reaching in press,'' in \emph{6th International Conference on
  Development and Learning and on Epigenetic Robotics}, 2016.

\bibitem{Law2011a}
J.~Law, M.~Lee, M.~H{\"u}lse, and A.~Tomassetti, ``The infant development
  timeline and its application to robot shaping,'' \emph{Adaptive Behavior},
  vol.~19, pp. 335--358, 2011.

\bibitem{Piek}
J.~P. Piek, \emph{Infant Motor Development}.\hskip 1em plus 0.5em minus
  0.4em\relax Human Kinetics, 2006.

\bibitem{gib:79}
J.~J. Gibson, \emph{The Ecological Approach to Visual Perception}.\hskip 1em
  plus 0.5em minus 0.4em\relax Boston: Houghton Mifflin, 1979.

\bibitem{spo:04}
O.~Sporns and T.~K. Pegors, ``Information--theoretical aspects of embodied
  artificial intelligence.'' in \emph{Embodied Artificial Intelligence},
  F.~Iida, R.~Pfeifer, L.~Steels, and Y.~Kuniyoshi, Eds.\hskip 1em plus 0.5em
  minus 0.4em\relax Berlin: Springer--Verlag, 2004, pp. 74--85.

\bibitem{kun:04}
Y.~Kuniyoshi, Y.~Yorozu, Y.~Ohmura, K.~Terada, T.~Otani, A.~Nagakubo, and
  T.~Yamamoto, ``From humanoid embodiment to theory of mind,'' in
  \emph{Embodied Artificial Intelligence}, F.~Iida, R.~Pfeifer, L.~Steels, and
  Y.~Kuniyoshi, Eds.\hskip 1em plus 0.5em minus 0.4em\relax Berlin:
  Springer--Verlag, 2004, pp. 202--218.

\bibitem{Lungarella2003}
M.~Lungarella, G.~Metta, R.~Pfeifer, and G.~Sandini, ``Developmental robotics:
  a survey,'' \emph{Connection Science}, vol.~15, pp. 151--190, 2003.

\bibitem{Asada2009}
M.~Asada, K.~Hosoda, Y.~Kuniyoshi, H.~Ishiguro, T.~Inui, Y.~Yoshikawa,
  M.~Ogino, and C.~Yoshida, ``Cognitive developmental robotics: A survey,''
  \emph{IEEE Transaction on Autonomous Mental Development}, vol.~1, no.~1,
  2009.

\bibitem{Gomez2004a}
G.~Gomez, M.~Lungarella, P.~E. Hotz, K.~Matsushita, and R.~Pfeifer,
  ``Simulating development in a real robot: on the concurrent increase of
  sensory, motor, and neural complexity,'' in \emph{Fourth International
  Workshop on Epigenetic Robotics}, M.~C.~D. in~Robotic~Systems, Ed., 2004.

\bibitem{Caligiore2008}
D.~Caligiore, T.~Ferrauto, D.~Parisi, N.~Accornero, M.~Capozza, and
  G.~Baldassarre, ``Using motor babbling and hebb rules for modeling the
  development of reaching with obstacles and grasping,'' in \emph{International
  Conference on Cognitive Systems}, 2008.

\bibitem{Asuni2005}
G.~Asuni, G.~Teti, and et~al., ``A bio-inspired sensory-motor neural model for
  neuro-robotic manipulation platform,'' in \emph{International Conference on
  Advanced Robotics,ICAR' 05. Proceedings., 12th}, Asuni2005, Ed., 2005.

\bibitem{Demiris2005}
\BIBentryALTinterwordspacing
Y.~Demiris and A.~Dearden, ``From motor babbling to hierarchical learning by
  imitation: a robot developmental pathway,'' pp. 31--37, 2005. [Online].
  Available: \url{http://cogprints.org/4961/}
\BIBentrySTDinterwordspacing

\bibitem{Lee2007a}
M.~H. Lee, Q.~Meng, and F.~Chao, ``Staged competence learning in developmental
  robotics,'' \emph{Adaptive Behavior}, vol.~15, pp. 241--255, 2007.

\bibitem{Lee2007b}
------, ``Developmental learning for autonomous robots,'' \emph{Robotics and
  Autonomous Systems}, vol.~55, no.~9, pp. 750--759, 2007.

\bibitem{Saegusa2008a}
R.~Saegusa, S.~Sakka, G.~Metta, and G.~Sandini, ``Autonomous learning
  evaluation toward active motor babbling,'' in \emph{2008 IEEE/RSJ
  International Conference on Intelligent Robots and Systems (IROS2008)
  Workshop: From motor to interaction learning in robots}, IEEE, Ed., 2008.

\bibitem{Saegusa2009IEEE/RSJ}
R.~Saegusa, G.~Metta, and G.~Sandini, ``Active learning for multiple
  sensorimotor coordination based on state confidence,'' in \emph{IEEE/RSJ
  international conference on Intelligent robots and systems}, 2009.

\bibitem{Saegusa2009HSI}
R.~Saegusa, G.~Metta, G.~Sandini, and S.~Sakka, ``Active learning for
  sensorimotor coordinations of autonomous robots,'' in \emph{2nd Conference on
  Human System Interactions. HSI '09.}, IEEE, Ed., 2009.

\bibitem{Laschi2008}
C.~Laschi, G.~Asuni, E.~Guglielmelli, G.~Teti, R.~Johansson, H.~Konosu,
  Z.~Wasik, M.~C. Carrozza, and P.~Dario, ``A bio-inspired predictive
  sensory-motor coordination scheme for robot reaching and preshaping,''
  \emph{Auton Robot}, vol.~25, pp. 85--101, 2008.

\bibitem{Rolf2010a}
M.~Rolf, J.~J. Steil, and M.~Gienger, ``Goal babbling permits direct leaning of
  inverse kinematics,'' in \emph{IEEE Transaction on Autonomous Mental
  Development}, 2010.

\bibitem{Rolf2010b}
------, ``Mastering growth while bootstrapping sensorimotor coordination,'' in
  \emph{International Conference on Epigenetic Robotics}, 2010.

\bibitem{Dewolf2011}
T.~DeWolf and C.~Eliasmith, ``The neural optimal control hierarchy for motor
  control,'' \emph{J. Neural Eng}, vol.~8, 2011.

\bibitem{Law2014b}
J.~Law, P.~Shaw, M.~Lee, and M.~Sheldon, ``From saccades to grasping: A model
  of coordinated reaching through simulated development on a humanoid robot,''
  \emph{IEEE Transaction on Autonomous Mental Development}, vol.~6, no.~2,
  2014.

\bibitem{Caligiore2014}
D.~Caligiore, D.~Parisi, and G.~Baldassarre, ``Integrating reinforcement
  learning, equilibrium points and minimum variance to understand the
  development of reaching: A computational model,'' \emph{Psychological
  Review}, vol. 121, no.~3, pp. 389--421, 2014.

\bibitem{autoencoder2006}
G.~E. Hinton and R.~R. Salakhutdinov, ``Reducing the dimensionality of data
  with neural networks,'' \emph{SCIENCE}, vol. VOL 313, 2006.

\bibitem{CCNbook}
R.~C. O'Reilly, Y.~Munakata, M.~J. Frank, and T.~E. Hazy, \emph{Computational
  Cognitive Neuroscience}, 1st~ed.\hskip 1em plus 0.5em minus 0.4em\relax Wiki
  Book, 2012.

\bibitem{Sucan2014}
I.~A. Sucan and S.~Chitta, \emph{MoveIt!}, 2014.

\end{thebibliography}

%

%








\end{document}